\documentclass{article}

\PassOptionsToPackage{square, numbers, compress}{natbib}

\usepackage[preprint]{neurips_2024}

\usepackage[utf8]{inputenc} % allow utf-8 input
\usepackage[T1]{fontenc}    % use 8-bit T1 fonts
\usepackage{hyperref}       % hyperlinks
\usepackage{url}            % simple URL typesetting
\usepackage{booktabs}       % professional-quality tables
\usepackage{amsfonts}       % blackboard math symbols
\usepackage{nicefrac}       % compact symbols for 1/2, etc.
\usepackage{microtype}      % microtypography
\usepackage[dvipsnames]{xcolor}
\usepackage{graphicx}
\usepackage{subcaption}
\usepackage{multirow}
\usepackage{Definitions}
\usepackage{commands}
\usepackage{wrapfig}
\usepackage{hyperref}
\usepackage{xspace}
\usepackage{color}
\usepackage{threeparttable}
\usepackage[toc]{appendix}

\newcommand{\update}[1]{{\textcolor{black}{#1}}}
\newcommand{\boldres}[1]{{\textbf{\textcolor{red}{#1}}}}
\newcommand{\secondres}[1]{{\underline{\textcolor{blue}{#1}}}}

\title{UniTST: Effectively Modeling Inter-Series and Intra-Series Dependencies for Multivariate Time Series Forecasting}
\author{%
    Juncheng Liu$^{1}$\thanks{Correspondence to: Juncheng Liu, Chenghao Liu <\{juncheng.liu, chenghao.liu\}@salesforce.com>} \quad Chenghao Liu$^{1*}$ \quad  Gerald Woo$^{1}$ \quad  Yiwei Wang$^{2}$ \quad  Bryan Hooi$^{3}$ \\
    \textbf{Caiming Xiong}$^{1}$ \quad \textbf{Doyen Sahoo}$^{1}$ \\
    Salesforce$^{1}$ \quad University of California, Los Angeles$^{2}$ \quad National University of Singapore$^{3}$ \\ 
}

\begin{document}

\maketitle

\begin{abstract}
Transformer-based models have emerged as powerful tools for multivariate time series forecasting (MTSF).
% attracting significant attention due to their ability to capture either temporal dynamics or channel dependencies. 
However, existing Transformer models often fall short of capturing both intricate dependencies across variate and temporal dimensions in MTS data. Some recent models are proposed to separately capture variate and temporal dependencies through either two sequential or parallel attention mechanisms. However, these methods cannot directly and explicitly learn the intricate inter-series and intra-series dependencies.
% , which makes them less effective in real applications. 
In this work, we first demonstrate that these dependencies are very important as they usually exist in real-world data. To directly model these dependencies, we propose a transformer-based model UniTST containing a unified attention mechanism on the flattened patch tokens. Additionally, we add a dispatcher module which reduces the complexity and makes the model feasible for a potentially large number of variates.
% Although our proposed model employs a simple architecture, it offers compelling performance in empirical experiments. 
% We demonstrate the effectiveness of our approach through extensive experiments on several datasets for both long-term and short-term forecasting, showing that our approach generally outperforms state-of-the-art methods. 
Although our proposed model employs a simple architecture, it offers compelling performance as shown in our extensive experiments on several datasets for time series forecasting. 
% Our study emphasizes the necessity and effectiveness of simultaneously capturing temporal and channel dependencies in MTSF, and our proposed designs represent a step toward this goal. 

% In this work, we address this limitation by proposing a simple yet effective model: the Cross-Channel, Cross-Temporal Time Series Transformer (C3TST) \TODO{Model Name: TBD}. Our model is designed to explicitly capture the interplay between channels and temporal dynamics, crucial for accurate forecasting in multivariate time series data.
% We demonstrate the effectiveness of our approach through extensive experiments on several datasets for both long-term and short-term forecasting. 
% Our results validate the superior performance of our C3TST compared with prior Transformer-based models, underscoring the importance of capturing cross-channel, cross-temporal dependencies for multivariate time series forecasting tasks. 
% In addition to achieving state-of-the-art forecasting accuracy, our model offers simplicity and efficiency, making it a compelling choice for real-world applications.
% Our study emphasizes the necessity of adaptively capturing both temporal and channel dependencies in multivariate time series forecasting, and our proposed C3TST model represents a step toward this goal.
\end{abstract}

\section{Introduction}

Inspired by success of Transformer-based models on various field such as natural language processing \citep{touvron2023llama1,vicuna2023,falcon40b,MosaicML2023Introducing,touvron2023llama,openaichatgptblog,bardclaudeblog,touvron2023llama} and computer vision \citep{wu2020visual,liu2021swin,jamil2023comprehensive}, 
Transformers have also garnered much attention in the community of multivariate time series forecasting (MTSF)~\citep{PatchTST,iTransformer, Autoformer, Crossformer, fedformer, carlini2023aligned,han2024softs}. % Pioneering transformer-based models employ linear or convolutional layers as embedding techniques to aggregate information at the same time step across different channels, and model dependencies along the temporal dimension via attention mechanisms~\citep{Informer,Autoformer,fedformer}. 
% However, it was pointed out that these models are often less effective than linear models~\citep{DLinear, han2023capacity} since such channel mixing structures are vulnerable to the distribution shift~\citep{DLinear}. 
Pioneering works~\citep{Informer,Autoformer,fedformer} treat multiple variates (aka channels) at each time step as the input unit for transformers, similar to tokens in the language domain, but its performance was even inferior to linear models~\citep{DLinear, han2023capacity}.
% Channel-independent methods~\citep{PatchTST, wang2023st} are subsequently proposed and achieve positive results by avoiding mixing noises from noisy channels. 
% Nevertheless, these channel-independent methods neglect the correlation between channels, which makes them difficult to further improve model performance.
Considering the noisy information from individual time points, \textit{Variate-Independent} and \textit{Patch-Based}~\citep{PatchTST} methods are subsequently proposed and achieve positive results by avoiding mixing noises from multiple variates and aggregating information from several adjacent time points as input. 
Nevertheless, these methods neglect the cross-variate relationships and interfere with the learning of temporal dynamics across variates.

\begin{figure}[th]
    \centering
    \includegraphics[width=\linewidth]{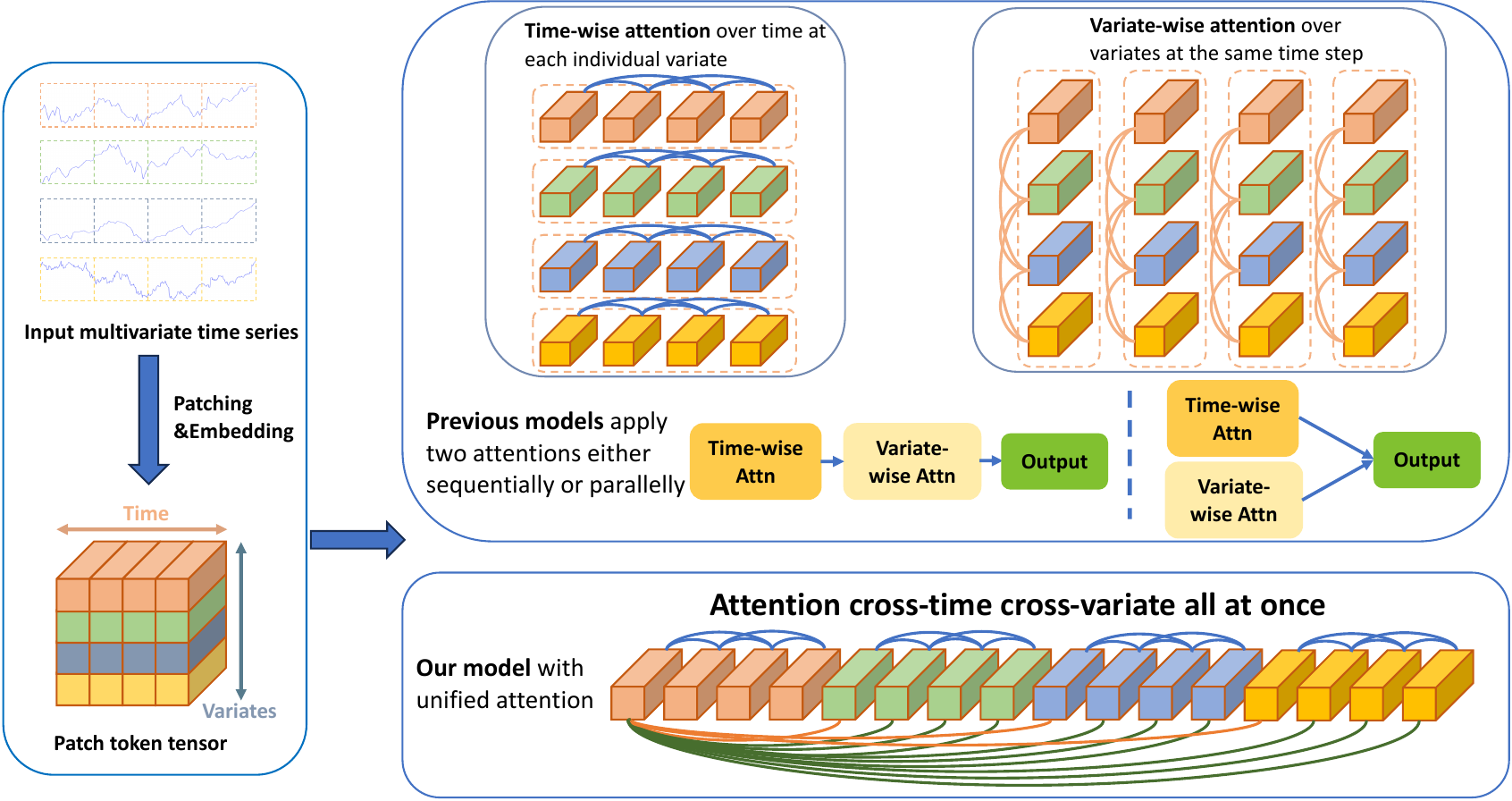}
    \caption{Comparison between our model and previous models. Previous models apply time-wise attention and variate-wise attention modules either sequentially or parallelly, which cannot capture cross-time cross-variate dependencies (i.e., {\color{ForestGreen} green links}) simultaneously like our model. 
    % \TODO{maybe rewrite the patch dimension in the cube to time?}
    % \TODO{Change the text on previous model - sequential or paralle}
    }
    \label{fig: comparison}
    \vspace{-5mm}
\end{figure}
% Recently, several works~\citep{iTransformer, Crossformer, carlini2023aligned, han2024softs, yu2024revitalizing, Timesnet} have been proposed to capture channel dependencies through mechanisms like attention.
% However, these models either only model channel dependencies with attention mechanisms on channel tokens~\citep{iTransformer}, or propose an attention mechanism with two separate stages to sequentially model time and channel dependencies~\citep{Crossformer, carlini2023aligned}, which are not direct and effective to capture time and channel dependencies simultaneously. 
% Specifically, iTransformer~\citep{iTransformer} embeds the whole time series of a channel into a token and utilizes "channel-wise self-attention" to model channel dependencies, which lacks direct modeling of intra-channel temporal dependencies.  
To tackle this problem, iTransformer~\citep{iTransformer} embeds the entire time series of a variate into a token and employs "variate-wise attention" to model variate dependencies. However, it lacks the capability to model intra-variate temporal dependencies.
% Two-stage methods~\citep{Crossformer,carlini2023aligned} generally apply two different attention mechanisms sequentially on time and channel dimensions of 2D input arrays to capture cross-time and cross-channel dependencies.
% Besides, \citet{yu2024revitalizing} propose to use parallel two-stage attention mechanisms to model inter-channel and intra-channel dependencies.  
Concurrently, several approaches ~\citep{Crossformer,carlini2023aligned,yu2024revitalizing} utilize both variate-wise attention and time(patch)-wise attention to capture inter-variate and intra-variate dependencies, either sequentially or parallelly.
% Specifically, two works \citep{Crossformer, carlini2023aligned} sequentially apply time-wise attention module and variate-wise attention module on temporal and variate dimensions of 2D input. 
% Another approach \citep{yu2024revitalizing} uses these two attention modules to model inter-variate and intra-variate dependencies in parallel.
Yet, they may raise the difficulty of modeling the diverse time and variate dependencies as the errors from one stage can affect the other stage and eventually the overall performance. 

Additionally, either two parallel or sequential attention mechanisms cannot explicitly model the direct dependencies across different variates and different times, which we show in Figure~\ref{fig: comparison}. Regardless of how previous works apply time-wise attention and variate-wise attention parallelly or sequentially, they would still lack the {\color{ForestGreen} green links} to capture cross-time cross-variate dependencies (aka inter-series intra-series dependencies) simultaneously as in our model. 
\begin{wrapfigure}{r}{0.4\textwidth}
   	\centering
	\includegraphics[width=0.4\textwidth]{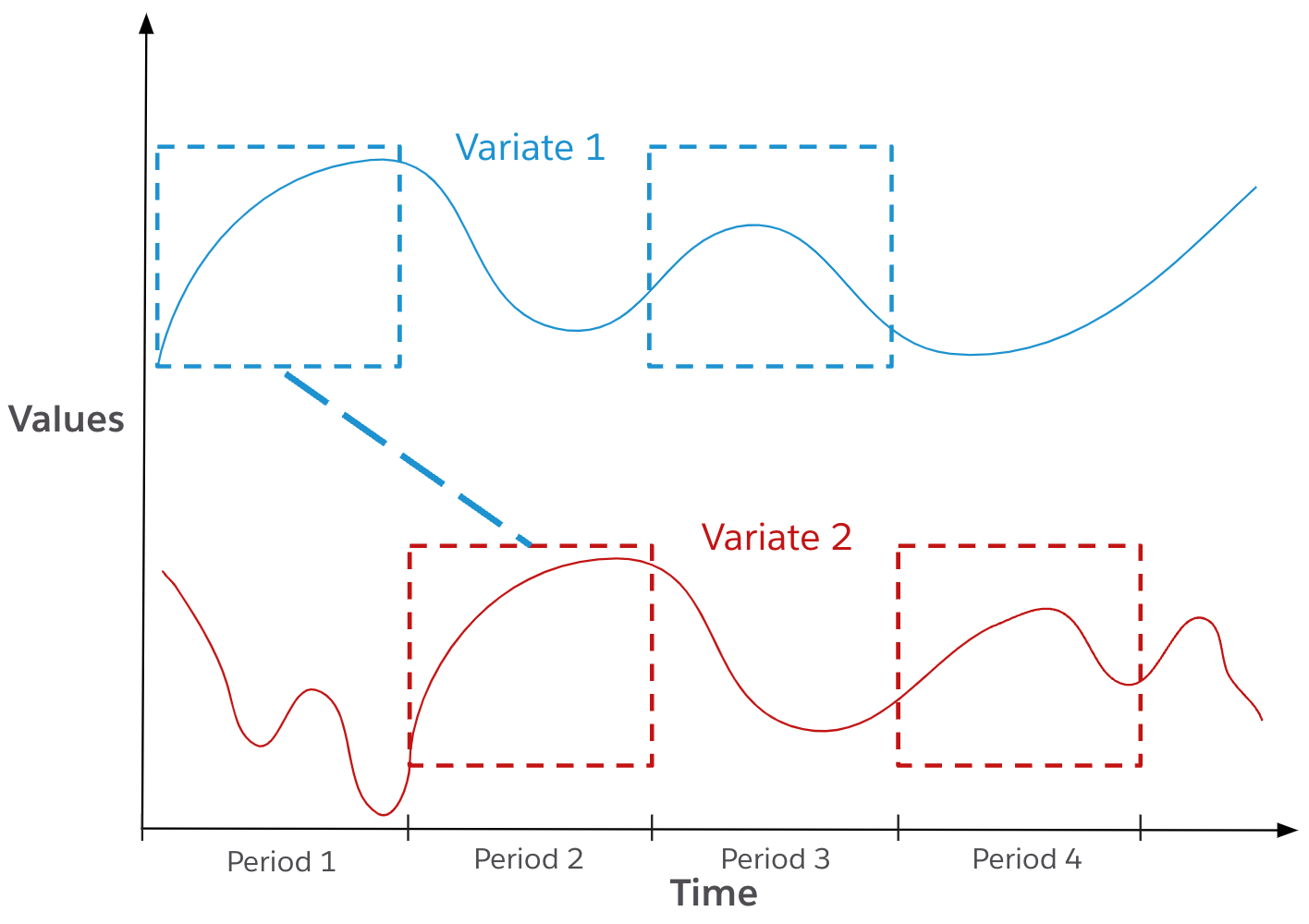}
	\caption{Explicit correlation between two sub-series at different periods from two different variates (i.e., strong correlation between period 1 of variate 1 and period 2 and variate 2).}
	\label{fig: explicit_corr}
% 	\caption{Averaged accuracies with respect to the length of chains.}
    \vspace{-5.0mm}
\end{wrapfigure}
To further explain, as we illustrate in Figure~\ref{fig: explicit_corr}, the time series of variate 1 during period 1 shares the same trend with the time series of variate 2 during period 2. This type of correlations cannot be directly modeled by previous works as it requires directly modeling cross-time cross-variate dependencies simultaneously. This type of correlation is important as it generally exists in real-world data as we further demonstrate in Sec \ref{sec: motivation}.
To mitigate the limitations of previous works, in this paper, we revisit the structure of multivariate time series transformers and propose a time series transformer with unified attention (\textit{\ourmodel}) as a fundamental backbone for multivariate forecasting. Technically, we flatten all patches from different variates into a unified sequence and adopt the attention for inter-variate and intra-variate dependencies simultaneously. To mitigate the high memory cost associated with the flattening strategy, we further develop a dispatcher mechanism to reduce complexity from quadratic to linear. 
% Despite the simple designs used in \ourmodel, empirical results show that \ourmodel achieves state-of-the-art performance on real-world benchmarks as shown in Section \ref{sec: exp_results}.
% in Figure \ref{fig: overall_performance}. 
% \begin{wrapfigure}{r}{0.3\textwidth}
%    	\centering
% 	\includegraphics[width=0.3\textwidth]{Figs/overall_performance.pdf}
% 	\caption{Performance of \ourmodel. \TODO{To revise and add more baselines}}
% 	\label{fig: overall_performance}
% % 	\caption{Averaged accuracies with respect to the length of chains.}
%     \vspace{-2.0mm}
% \end{wrapfigure}
% \TODO{to re-emphasize the main point of our work, not mainly about a model but providing a new view for researcher to think about what design we should have for modeling relationship across temporal and channel dimensions}. We believe that the principle of our design can 
Our contributions are summarized as follows: 
\setlist{leftmargin=*}
\begin{itemize}
% \item We discuss the transformer structure for multivariate time series and argue that the ability to model both inter-variate and intra-variate dependencies is crucial in multivariate time series forecasting. 
% Through detailed analysis, we identify that current sequential and parallel approaches result in suboptimal performance, thereby limiting the predictive capabilities of Transformers.
\item We point out the limitation of previous transformer models for multivariate time series forecasting: their lack of ability to simultaneously capture both inter-variate and intra-variate dependencies. With evidence in real-world data, we demonstrate that these dependencies are important and commonly exist. 

% \item We propose a simple yet effective transformer-based model (\ourmodel) with a unified attention mechanism and a dispatcher module that tokenizes and flatten all patches from different channels to directly capture diverse dependencies across different channels and temporal dimensions simultaneously. 
\item To mitigate the limitation,  we propose \ourmodel as a simple, general yet effective transformer for modeling multivariate time series data, which flattens all patches from different variates into a unified sequence to effectively capture inter-variate and intra-variate dependencies. 
\item Despite the simple designs used in \ourmodel, we empirically demonstrate that \ourmodel achieves state-of-the-art performance on real-world benchmarks for both long-term and short-term forecasting with improvements up to 13\%. In addition, we provide results of the ablation study and visualization to further demonstrate the effectiveness of our model. 
\end{itemize}
% Paragraph 5: Introduce our model with detailed components. 1) Attentions on different patches cross different channels; 2) mentioned the memory cost incurred if we directly have this design, then we propose a new router mechanism to mitigate this memory issues. 

% Paragraph 6: Provide a figure like Figure 1 in iTransformer to show that empirically our model works well although it's simple. Furthermore, mentions how our model can bring new views on this field and try to highlight something more valuable than the simple architecture. Then list out the contributions. 

\section{Related Work}
% To have a similar figure as Figure 3 of iTransformer paper. Then put our model into the category (modified architecture and no modified component).

% \subsection{Transformer for Multivariate Time Series Forecasting}
% 1) channel-mixing transformer: embed the temporal tokens -> contain values of multivariate at the same time step., It's not effective as the points at the same time steps may have different physical meaning records. They don't share the same semantic meaning. And it's not robust against the distribution shift, which makes them worse than linear models. 
% 2) Channel-independent: PatchTST 
% 3) Talked more on the recent works which try to capture cross-channel dependencies. 
Recently, many Transformer-based models have been also proposed for multivariate time series forecasting and demonstrated great potential~\citep{Pyraformer, Autoformer,Informer,Crossformer,fedformer, LogSparse}. 
Several approaches~\citep{Autoformer, Informer, fedformer} embed temporal tokens that contain the multivariate representation of each time step and utilize attention mechanisms to model temporal dependencies. However, due to the vulnerability to the distribution shift, these models with such channel mixing structure are often outperformed by simple linear models~\citep{DLinear,han2023capacity}. 
Subsequently, PatchTST~\citep{PatchTST} considers channel independence and models temporal dependencies within each channel to make predictions independently. 
% It achieves better performance by avoiding mixing noises from noisy channels. 
Nonetheless, it ignores the correlation between variates, which may hinder its performance. 

To model variate dependencies, in the past two years, several works have been proposed~\citep{iTransformer, Crossformer, carlini2023aligned, han2024softs, yu2024revitalizing, Timesnet}. iTransformer~\citep{iTransformer} models channel dependencies by embedding the whole time series of a variate into a token and using "variate-wise attention". 
Crossformer~\citep{Crossformer} uses the encoder-decoder architecture with two-stage attention layers to sequentially model cross-time dependencies and then cross-variate dependencies. 
CARD~\citep{carlini2023aligned} employs the encoder-only architecture utilizing a similar sequential two-stage attention mechanism for cross-time, cross-channel dependencies and a token blend module to capture multi-scale information. 
Leddam~\citep{yu2024revitalizing} designs a learnable decomposition and a dual attention module that parallelly model inter-variate dependencies with "channel-wise attention" and intra-variate temporal dependencies with "auto-regressive attention".
In summary, these works generally model intra-variate and inter-variate dependencies separately (either sequentially or parallelly), and aggregate these two types of information to get the outputs. 
In contrast, our model has a general ability to directly capture inter-variate and intra-variate dependencies simultaneously, which is more effective. We provide more discussion on the comparison between our model and previous models in Section~\ref{sec: more_discussion_models}.

\section{Preliminary and Motivation}
\label{sec: motivation}
In multivariate time series forecasting, given historical observations $\mathbf{X}_{:, t:t+L} \in \mathbb{R}^{N \times L}$ with $L$ time steps and $N$ variates, the task is to predict the future $S$ time steps, i.e., $\mathbf{X}_{:, t+L+1:t+L+S} \in \mathbb{R}^{N \times S}$. For convenience, we denote $\mathbf{X}_{i, :} = \mathbf{x}^{(i)}$ as the whole time series of the $i$-th variate and $\mathbf{X}_{:, t}$ as the recorded time points of all variates at time step $t$.

%A multivariate time series (MTS) is defined as $\mathbf{X} = \{\mathbf{x}^{(1)}, \mathbf{x}^{(2)}, ..., \mathbf{x}^{(N)}\} \in \mathbb{R}^{N\times T}$, where $N$ is the number of variates (also called channels), $T$ is the number of time steps. We denote $\mathbf{X}_{i, :} = \mathbf{x}^{(i)}$ as the whole time series of the $i$-th variate and $\mathbf{X}_{:, t}$ as the recorded time points at time step $t$. 
%Given historical observations $\mathbf{X}_{:, t:t+L} \in \mathbb{R}^{N \times L}$ with $L$ time steps, the MTS task requires a model to predict future consecutive $S$ time steps, i.e., $\mathbf{X}_{:, t+L+1:t+S} \in \mathbb{R}^{N \times S}$.  

To illustrate the diverse cross-time and cross-variate dependencies from real-world data, we use the following correlation coefficient between $\mathbf{x}^{(i)}_{t:t+L}$ and $\mathbf{x}^{(j)}_{t+L:t+2L}$ to measure it,
\begin{definition}[Cross-Time Cross-Variate Correlation Coefficient]
\begin{equation}\label{eq:cross corr}
    R^{(i,j)}(t,t',L) = \frac{\operatorname{Cov}(\mathbf{x}^{(i)}_{t:t+L}, \mathbf{x}^{(j)}_{t':t'+L})}{\sigma^{(i)}\sigma^{(j)}}  = \frac{1}{L}\sum_{k=0}^{L}{\frac{\mathbf{x}^{(i)}_{t+k}-\mu^{(i)}}{\sigma^{(i)}}\cdot\frac{\mathbf{x}^{(j)}_{t'+k}-\mu^{(j)}}{\sigma^{(j)}}},
\end{equation}
where $\mu^{(\cdot)}$ and $\sigma^{(\cdot)}$ are the mean and standard deviation of corresponding time series patches. 
\end{definition}
\begin{wrapfigure}{r}{0.35\textwidth}
    \vspace{-3.0mm}
   \centering
	\includegraphics[width=0.35\textwidth]{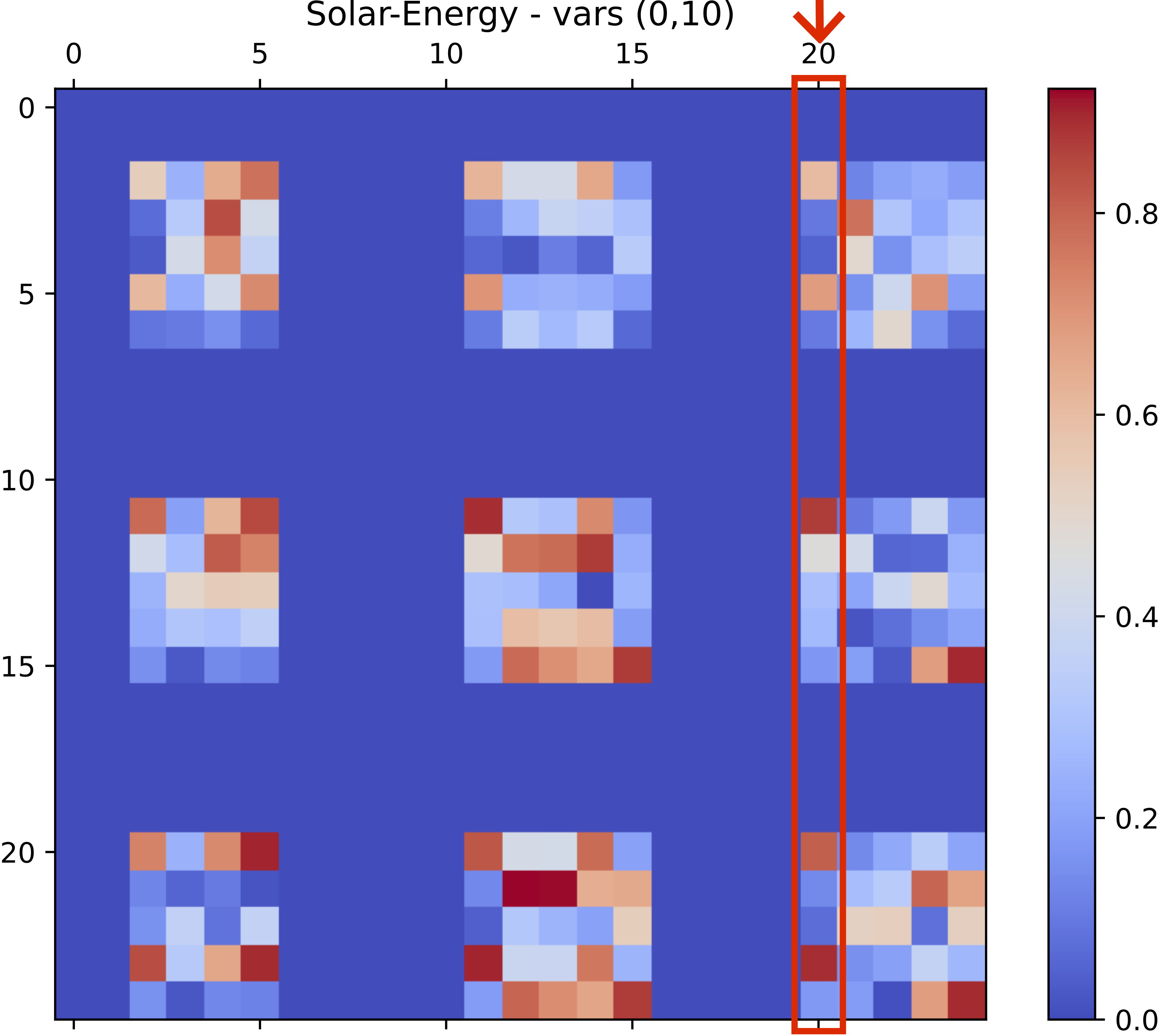}
	\caption{Correlation between patches from different variates. x-axis: patch indices in variate 10, y-axis: patch indices in variate 0.}
 %    \includegraphics[width=0.3\textwidth]{Figs/weather-vars(0,14).pdf}
	% \caption{Correlation between different patches from variate 0 and variate 14 on Weather dataset.}
	\label{fig: motivation_real_data}
% 	\caption{Averaged accuracies with respect to the length of chains.}
    \vspace{-2.0mm}
\end{wrapfigure}
Utilizing the above correlation coefficient, we can quantify and further understand the diverse cross-time cross-variate correlation. We visualize the correlation coefficient between different time periods from two different variates in Figure \ref{fig: motivation_real_data}.
We split the time series into several patches and each patch denotes a time period containing 16 time steps. 
In Figure \ref{fig: motivation_real_data}, we can see that, first, given a pair of variates, the inter-variate dependencies are quite different for different patches.
Looking at the column of Patch 20 in variate 10, it is strongly correlated with patch 3, 5, 11, 20, 24 of variate 0, while it is very weakly correlated with all other patches from variate 0.  
It suggests that there is no consistent correlation pattern for different patch pairs of two variates (i.e., not all the same coefficient at a row/column in the correlation map) and inter-variate dependencies are actually at the fine-grained patch level. 
Therefore, previous transformer-based models have a deficiency in directly capturing this kind of dependencies. The reason is that they either only capture the dependencies for the whole time series between two variates without considering the fine-grained temporal dependencies across different variates \citep{iTransformer} or use two separate attention mechanisms \citep{Crossformer, carlini2023aligned, yu2024revitalizing} which are indirect and unable to explicitly learn these dependencies. 
In Appendix~\ref{app:examples_diverse_dependencies}, we provide more examples to demonstrate the ubiquity and the diversity of these cross-time cross-variate correlations. 
% inter-variate dependencies are at the fine-grained patch level, which may not be captured by methods 
% \TODO{add a few more plots in other datasets in appendix, to say that these dependencies commonly exist in the real-world data.} 

% Previous Transformer-based models have a deficiency in directly capturing this kind of dependencies. They either only capture the dependencies for the whole time series between two variates without considering the fine-grained temporal dependencies across different channels \citep{iTransformer} or use two separate attention mechanisms \citep{Crossformer, carlini2023aligned, yu2024revitalizing} which are indirect and unable to explicitly learn these dependencies. 
% \TODO{May change this sentence} Therefore, in this work, we aim to propose a model with the ability to explicitly directly capture cross-time cross-variate interactions for multivariate data. 
Motivated by the deficiency of previous models in capturing these important dependencies, in this work, we aim to propose a model with the ability to explicitly directly capture cross-time cross-variate interactions for multivariate data. 

\begin{figure}[th]
    \centering
    \includegraphics[width=\linewidth]{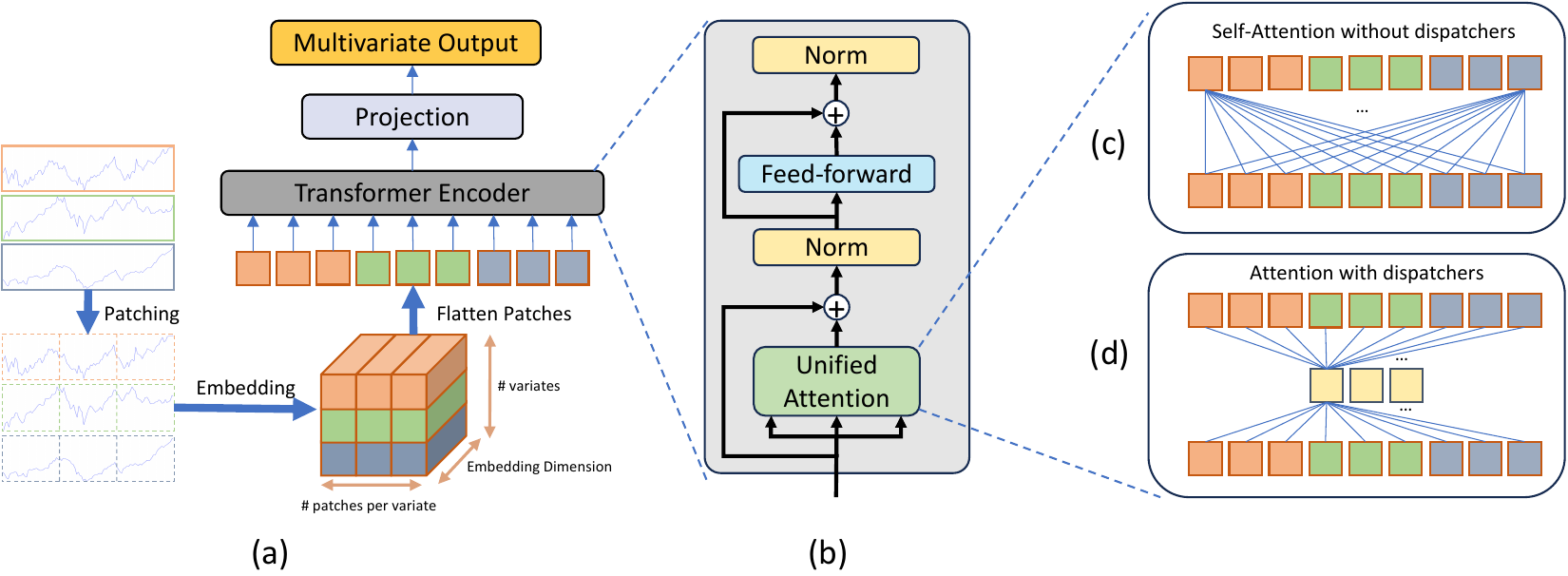}
    \caption{Framework Overview. We flatten the patches from all variates into a sequence as the input of the Transformer Encoder and replace the original self-attention with the proposed unified attention with dispatchers to reduce the memory complexity.}
    \label{fig: framework}
\end{figure}
\section{Methodology}
In this section, we describe our proposed Transformer-based method (\ourmodel) for modeling inter-variate and intra-variate dependencies for multivariate time series forecasting. Then, we discuss and compare our model with previous Transformer-based models in detail.   

\subsection{Model Structure Overview}
\label{sec: model_structure}
We illustrate our proposed \ourmodel with a unified attention mechanism in Figure \ref{fig: framework}. 
% \TODO{Add more descriptions here.}
\paragraph{Embedding the patches from different variates as the tokens} 
Given the time series with $N$ variates $X \in \mathbb{R}^{N \times T}$, we divide each univariate time series $x^{i}$ into patches as in~\citet{PatchTST, Crossformer}. With the patch length $l$ and the stride $s$, for each variate $i$, we obtain a patch sequence $x_p^{i} \in \mathbb{R}^{p\times l}$ where $p$ is the number of patches. Considering all variates, the tensor containing all patches is denoted as $X_p \in \mathbb{R}^{N \times p\times l}$, where $N$ is the number of variates. 
With each patch as a token, the 2D token embeddings are generated using a linear projection with position embeddings: 
\begin{equation}
    H = \text{Embedding}(X_p) = X_pW + W_{pos} \in \mathbb{R}^{N\times p \times d},
\end{equation}
where $W \in \mathbb{R}^{l \times d}$ is the learnable projection matrix and $W_{pos} \in \mathbb{R}^{N \times p \times d}$ is the learnable position embeddings. With 2D token embeddings, we denote $H^{(i,k)}$ is the token embedding of the $k$-th patches in the $i$-th variate, resulting in $N \times p$ tokens. 

\paragraph{Self attention on the flattened patch sequence}
Considering any two tokens, there are two relationships: 1) they are from the same variate; 2) they are from two different variates. These represent intra-variate and cross-variate dependencies, respectively. 
A desired model should have the ability to capture both types of dependencies, especially cross-variate dependencies. 
To capture both intra-variate and cross-variate dependencies among tokens, we flatten the 2D token embedding matrix $H$ into a 1D sequence with $N \times p$ tokens. 
We use this 1D sequence $X' \in \mathbb{R}^{(N\times p) \times d}$ as the input and feed it to a vanilla Transformer encoder. The multi-head self-attention (MSA) mechanism is directly applied to the 1D sequence:

\begin{equation}
    O = \text{MSA}(Q,K,V) = \text{Softmax}(\frac{QK^T}{\sqrt{d_k}})V,
\end{equation}
with the query matrix $Q=X'W_Q \in \mathbb{R}^{(N\times p) \times d_k}$, the key matrix $K=X'W_K \in \mathbb{R}^{(N\times p) \times d_k}$, the value matrix $V=X'W_V \in \mathbb{R}^{(N\times p) \times d}$, and $W_Q, W_K \in \mathbb{R}^{d \times d_k}$, $W_V \in \mathbb{R}^{d \times d}$. 
The MSA helps the model to capture dependencies among all tokens, including both intra-variate and cross-variate dependencies. However, the MSA results in an attention map with the memory complexity of $O(N^2 p^2)$, which is very costly when we have a large number of variates $N$. 

\paragraph{Dispatchers}
In order to mitigate the complexity of possible large $N$, we further propose a dispatcher mechanism to aggregate and dispatch the dependencies among tokens. We add $k (k << N) $ learnable embeddings as dispatchers and use cross attention to distribute the dependencies. The dispatchers aggregate the information from all tokens by using the dispatcher embeddings $D$ as the query and the token embeddings as the key and value:
\begin{equation}
    D' = \text{Attention}(DW_{Q_1}, X'W_{K_1}, X'W_{V_1}) = \text{Softmax}(\frac{DW_{Q_1}(X'W_{K_1})^T}{\sqrt{d_k}})X'W_{V_1},
\end{equation}
where the complexity is $O(kNp)$, and $W_{Q_1}, W_{K_1} \in \mathbb{R}^{d\times d_k}, W_{V_1} \in \mathbb{R}^{d\times d}$. 
After that, the dispatchers distribute the dependencies information to all tokens by setting the token embeddings as the key and the updated dispatcher embeddings $D'$ as the key and value:
\begin{equation}
    O' = \text{Attention}(X'W_{Q_2}, D'W_{K_2}, D'W_{V_2}) = \text{Softmax}(\frac{X'W_{Q_2}(D'W_{K_2})^T}{\sqrt{d_k}})D'W_{V_2},
\end{equation}
where the complexity is also $O(kNp)$. Therefore, the overall complexity of our dispatcher mechanism is $O(kNp)$, instead of $O(N^2 p^2)$ if we directly use self-attention on the flattened patch sequence. 
With the dispatcher mechanism, the dependencies between any two patches can be explicitly modeled through attention, no matter if they are from the same variate or different variates. 

In a transformer block, the output of attention $O'$ is passed to a BatchNorm Layer and a feedforward layer with residual connections. After stacking several layers, the token representations are generated as $Z^{N \times d'}$. In the end, a linear projection is used to generate the prediction $\hat{\textbf{X}} \in \mathbb{R}^{N \times S}$. 

\paragraph{Loss function}
The Mean-Squared Error (MSE) loss is used as the objective function to measure the difference between the ground truth and the generated predictions: $\mathcal{L} = \frac{1}{NS}\sum_{i}^{N}(\hat{\textbf{X}}^{(i)} - \textbf{X}_{i, t+L+1:t+S})^2$

\subsection{Discussion and Comparison with Previous Models}
\label{sec: more_discussion_models}
Our proposed model is an encoder-only transformer model containing a unified attention mechanism with dispatchers. The model explicitly learns both intra-variate and inter-variate temporal dependencies among different patch tokens through attention, which means that it can directly capture the correlation between two time series at different periods from different variates. 
In contrast, these dependencies cannot be directly and explicitly captured by previous works which claim that 
they model variate dependencies~\citep{iTransformer, Crossformer, carlini2023aligned, yu2024revitalizing}. For example, iTransformer~\citep{iTransformer} captures variate dependencies using the whole time series of a variate as a token. It loses the ability to capture the fine-grained temporal dependencies across channels or within a channel. 
Crossformer~\citep{Crossformer} and CARD~\citep{carlini2023aligned} both propose to use a sequential two-stage attention mechanism to first capture dependencies on time dimensions and then capture dependencies on variate dimensions. This sequential manner does not directly capture cross-time cross-variate dependencies simultaneously, which makes them less effective as shown in their empirical performance. 
In contrast, our proposed model uses a more unified attention on a flattened patch sequence with all patches from different channels, allowing direct and explicit modeling cross-time cross-variate dependencies. 
In addition, \citet{yu2024revitalizing} propose a dual attention module with an iTransformer-like encoder to inter-variate dependencies and an auto-regressive self-attention on each channel to capture intra-variate dependencies separately. In this way, it also cannot directly capture cross-variate temporal dependencies between two patch tokens at different time steps from different variates (e.g., $H^{(i,k)}$, while our model is able to directly capture these dependencies. 

Worth noting that our proposed model is a more general case to directly capture intra-variate and inter-variate dependencies at a more fine-grained level (i.e., patch level from different variates at different times). Moreover, our model employs simple architectures that can be easily implemented while the empirical results shows the effectiveness of our model in Section \ref{sec: exp_results}.
% Additionally, our proposed model enjoys simple architecture 
% In summary, our proposed model enjoys simple architectures with the ability to capture more intricate 

\section{Experiments}
We conduct comprehensive experiments to evaluate our proposed model \ourmodel and compare it with 11 representative baselines for both short-term and long-term time series forecasting on 13 datasets. Additionally, we further dive deeper into model analysis to examine the effectiveness of our model from different aspects. 

% \paragraph{Datasets}
% xxx

\subsection{Forecasting Results}
\label{sec: exp_results}
We conduct extensive experiments to compare our model with several representative time series models for both short-term and long-term time series forecasting. 
The detail of experimental setting and hyperparameter setting are discussed in Appendix~\ref{app: experimental_setting}

\paragraph{Baselines} We select 11 well-known forecasting models as our baselines, including (1) Transformer-based models: iTransformer \citep{iTransformer}, Crossformer \citep{Crossformer}, FEDformer~\citep{fedformer}, Stationary~\citep{Stationary}, PatchTST~\citep{PatchTST}; 
(2) Linear-based methods: DLinear~\citep{DLinear}, RLinear~\citep{li2023revisiting}, TiDE~\citep{das2023long}; (3) Temporal Convolutional Network (TCN)-based methods: TimesNet~\citep{Timesnet}, SCINet~\citep{SCINet}.
% ; and (4) MLP-based method: SOFTS~\citep{han2024softs}

\begin{table}[htbp]
  \caption{Multivariate long-term forecasting results with prediction lengths $S\in\{96, 192, 336, 720\}$ and fixed lookback length $T=96$. Results are averaged from all prediction lengths. Full results are listed in Appendix~\ref{app:full_reults}, Table~\ref{tab:full_baseline_results}.}
  % \vspace{-5pt}
  \label{tab:main_result_long_term}
  \renewcommand{\arraystretch}{0.85} 
  \centering
  \resizebox{1\columnwidth}{!}{
  \begin{threeparttable}
  \begin{small}
  \renewcommand{\multirowsetup}{\centering}
  \setlength{\tabcolsep}{1.45pt}
  \begin{tabular}{c|cc|cc|cc|cc|cc|cc|cc|cc|cc|cc|cc|cc}
    \toprule
    {\multirow{2}{*}{Models}} & 
    \multicolumn{2}{c}{\rotatebox{0}{\scalebox{0.75}{{\textbf{\ourmodel}}}}} &
    \multicolumn{2}{c}{\rotatebox{0}{\scalebox{0.75}{{iTransformer}}}} &
    \multicolumn{2}{c}{\rotatebox{0}{\scalebox{0.8}{\update{RLinear}}}} &
    \multicolumn{2}{c}{\rotatebox{0}{\scalebox{0.8}{PatchTST}}} &
    \multicolumn{2}{c}{\rotatebox{0}{\scalebox{0.8}{Crossformer}}} &
    \multicolumn{2}{c}{\rotatebox{0}{\scalebox{0.8}{TiDE}}} &
    \multicolumn{2}{c}{\rotatebox{0}{\scalebox{0.8}{{TimesNet}}}} &
    \multicolumn{2}{c}{\rotatebox{0}{\scalebox{0.8}{DLinear}}} &
    \multicolumn{2}{c}{\rotatebox{0}{\scalebox{0.8}{SCINet}}} &
    \multicolumn{2}{c}{\rotatebox{0}{\scalebox{0.8}{FEDformer}}} &
    \multicolumn{2}{c}{\rotatebox{0}{\scalebox{0.8}{Stationary}}} &
    \multicolumn{2}{c}{\rotatebox{0}{\scalebox{0.8}{Autoformer}}}  \\
     &
     
    \multicolumn{2}{c}{\scalebox{0.8}{\textbf{(Ours)}}} &
    \multicolumn{2}{c}{\scalebox{0.8}{\citeyearpar{iTransformer}}} & 
    \multicolumn{2}{c}{\scalebox{0.8}{\citeyearpar{li2023revisiting}}} & 
    \multicolumn{2}{c}{\scalebox{0.8}{\citeyearpar{PatchTST}}} & 
    \multicolumn{2}{c}{\scalebox{0.8}{\citeyearpar{Crossformer}}} & 
    \multicolumn{2}{c}{\scalebox{0.8}{\citeyearpar{das2023long}}} & 
    \multicolumn{2}{c}{\scalebox{0.8}{\citeyearpar{Timesnet}}} & 
    \multicolumn{2}{c}{\scalebox{0.8}{\citeyearpar{DLinear}}} &
    \multicolumn{2}{c}{\scalebox{0.8}{\citeyearpar{SCINet}}} & 
    \multicolumn{2}{c}{\scalebox{0.8}{\citeyearpar{fedformer}}} &
    \multicolumn{2}{c}{\scalebox{0.8}{\citeyearpar{Stationary}}} &
    \multicolumn{2}{c}{\scalebox{0.8}{\citeyearpar{Autoformer}}} \\
    \cmidrule(lr){2-3} \cmidrule(lr){4-5}\cmidrule(lr){6-7} \cmidrule(lr){8-9}\cmidrule(lr){10-11}\cmidrule(lr){12-13} \cmidrule(lr){14-15} \cmidrule(lr){16-17} \cmidrule(lr){18-19} \cmidrule(lr){20-21} \cmidrule(lr){22-23} \cmidrule(lr){24-25} 
    {Metric}  & \scalebox{0.8}{MSE} & \scalebox{0.8}{MAE}  & \scalebox{0.8}{MSE} & \scalebox{0.8}{MAE}  & \scalebox{0.8}{MSE} & \scalebox{0.8}{MAE}  & \scalebox{0.8}{MSE} & \scalebox{0.8}{MAE}  & \scalebox{0.8}{MSE} & \scalebox{0.8}{MAE}  & \scalebox{0.8}{MSE} & \scalebox{0.8}{MAE} & \scalebox{0.8}{MSE} & \scalebox{0.8}{MAE} & \scalebox{0.8}{MSE} & \scalebox{0.8}{MAE} & \scalebox{0.8}{MSE} & \scalebox{0.8}{MAE} & \scalebox{0.8}{MSE} & \scalebox{0.8}{MAE} & \scalebox{0.8}{MSE} & \scalebox{0.8}{MAE} & \scalebox{0.8}{MSE} & \scalebox{0.8}{MAE} \\
    \toprule
    \scalebox{0.95}{ECL}  & \boldres{\scalebox{0.8}{0.166}} & \boldres{\scalebox{0.8}{0.262}} & \secondres{\scalebox{0.8}{0.178}} & \secondres{\scalebox{0.8}{0.270}} &\scalebox{0.8}{0.219} &\scalebox{0.8}{0.298} & \scalebox{0.8}{0.205} & \scalebox{0.8}{0.290} & \scalebox{0.8}{0.244} & \scalebox{0.8}{0.334}  & \scalebox{0.8}{0.251} & \scalebox{0.8}{0.344} &\scalebox{0.8}{0.192} &\scalebox{0.8}{0.295} &\scalebox{0.8}{0.212} &\scalebox{0.8}{0.300} & \scalebox{0.8}{0.268} & \scalebox{0.8}{0.365} &\scalebox{0.8}{0.214} &\scalebox{0.8}{0.327} &{\scalebox{0.8}{0.193}} &{\scalebox{0.8}{0.296}} &\scalebox{0.8}{0.227} &\scalebox{0.8}{0.338} \\% &\scalebox{0.8}{0.311} &\scalebox{0.8}{0.397} \\
    \midrule
    \scalebox{0.95}{\update{ETTm1}} & \boldres{\scalebox{0.8}{0.379}} & \boldres{\scalebox{0.8}{0.394}} & \scalebox{0.78}{0.407} & \scalebox{0.78}{0.410} & \scalebox{0.78}{0.414} & \scalebox{0.78}{0.407} & \secondres{\scalebox{0.78}{0.387}} & \secondres{\scalebox{0.78}{0.400}} & \scalebox{0.78}{0.513} & \scalebox{0.78}{0.496} & \scalebox{0.78}{0.419} & \scalebox{0.78}{0.419} &{\scalebox{0.78}{0.400}} &{\scalebox{0.78}{0.406}}  &{\scalebox{0.78}{0.403}} &{\scalebox{0.78}{0.407}} & \scalebox{0.78}{0.485} & \scalebox{0.78}{0.481}  &\scalebox{0.78}{0.448} &\scalebox{0.78}{0.452} &\scalebox{0.78}{0.481} &\scalebox{0.78}{0.456} &\scalebox{0.78}{0.588} &\scalebox{0.78}{0.517} \\
    \midrule 
    \scalebox{0.95}{\update{ETTm2}} & \boldres{\scalebox{0.8}{0.280}} & \boldres{\scalebox{0.8}{0.326}} & {\scalebox{0.78}{0.288}} & {\scalebox{0.78}{0.332}} & \secondres{\scalebox{0.78}{0.286}} & \secondres{\scalebox{0.78}{0.327}} & {\scalebox{0.78}{0.281}} & \boldres{\scalebox{0.78}{0.326}} & \scalebox{0.78}{0.757} & \scalebox{0.78}{0.610} & \scalebox{0.78}{0.358} & \scalebox{0.78}{0.404} &{\scalebox{0.78}{0.291}} &{\scalebox{0.78}{0.333}} &\scalebox{0.78}{0.350} &\scalebox{0.78}{0.401} & \scalebox{0.78}{0.571} & \scalebox{0.78}{0.537} &\scalebox{0.78}{0.305} &\scalebox{0.78}{0.349} &\scalebox{0.78}{0.306} &\scalebox{0.78}{0.347} &\scalebox{0.78}{0.327} &\scalebox{0.78}{0.371} \\
    \midrule 
    \scalebox{0.95}{\update{ETTh1}} & \secondres{\scalebox{0.8}{0.442}} & \secondres{\scalebox{0.8}{0.435}} & {\scalebox{0.78}{0.454}} & {\scalebox{0.78}{0.447}} & {\scalebox{0.78}{0.446}} & \boldres{\scalebox{0.78}{0.434}} & \scalebox{0.78}{0.469} & \scalebox{0.78}{0.454} & \scalebox{0.78}{0.529} & \scalebox{0.78}{0.522} & \scalebox{0.78}{0.541} & \scalebox{0.78}{0.507} &\scalebox{0.78}{0.458} &{\scalebox{0.78}{0.450}} &{\scalebox{0.78}{0.456}} &{\scalebox{0.78}{0.452}} & \scalebox{0.78}{0.747} & \scalebox{0.78}{0.647} &\boldres{\scalebox{0.78}{0.440}} &\scalebox{0.78}{0.460} &\scalebox{0.78}{0.570} &\scalebox{0.78}{0.537} &\scalebox{0.78}{0.496} &\scalebox{0.78}{0.487} \\
    \midrule 
    \scalebox{0.95}{\update{ETTh2}} & \boldres{\scalebox{0.8}{0.363}} & \boldres{\scalebox{0.8}{0.393}} & {\scalebox{0.78}{0.383}} & {\scalebox{0.78}{0.407}} & \secondres{\scalebox{0.78}{0.374}} & \secondres{\scalebox{0.78}{0.398}} & {\scalebox{0.78}{0.387}} & {\scalebox{0.78}{0.407}} & \scalebox{0.78}{0.942} & \scalebox{0.78}{0.684} & \scalebox{0.78}{0.611} & \scalebox{0.78}{0.550}  &{\scalebox{0.78}{0.414}} &{\scalebox{0.78}{0.427}} &\scalebox{0.78}{0.559} &\scalebox{0.78}{0.515} & \scalebox{0.78}{0.954} & \scalebox{0.78}{0.723} &\scalebox{0.78}{{0.437}} &\scalebox{0.78}{{0.449}} &\scalebox{0.78}{0.526} &\scalebox{0.78}{0.516} &\scalebox{0.78}{0.450} &\scalebox{0.78}{0.459}  \\
    \midrule 
    \scalebox{0.95}{\update{Exchange}} & \boldres{\scalebox{0.8}{0.351}} & \boldres{\scalebox{0.8}{0.398}} & {\scalebox{0.8}{0.360}} & \secondres{\scalebox{0.8}{0.403}} &\scalebox{0.8}{0.378} &\scalebox{0.8}{0.417} & \scalebox{0.8}{0.367} & {\scalebox{0.8}{0.404}} & \scalebox{0.8}{0.940} & \scalebox{0.8}{0.707} & \scalebox{0.8}{0.370} & \scalebox{0.8}{0.413} &{\scalebox{0.8}{0.416}} &{\scalebox{0.8}{0.443}} & \secondres{\scalebox{0.8}{0.354}} &\scalebox{0.8}{0.414} & \scalebox{0.8}{0.750} & \scalebox{0.8}{0.626} &{\scalebox{0.8}{0.519}} &\scalebox{0.8}{0.429} &\scalebox{0.8}{0.461} &{\scalebox{0.8}{0.454}} &\scalebox{0.8}{0.613} &\scalebox{0.8}{0.539} \\

    \midrule 
    \scalebox{0.95}{Traffic} & \secondres{\scalebox{0.8}{0.439}} & \boldres{\scalebox{0.8}{0.274}} & \boldres{\scalebox{0.8}{0.428}} & \secondres{\scalebox{0.8}{0.282}} &\scalebox{0.8}{0.626} &\scalebox{0.8}{0.378} & {\scalebox{0.8}{0.481}} & {\scalebox{0.8}{0.304}} & \scalebox{0.8}{0.550} & {\scalebox{0.8}{0.304}} & \scalebox{0.8}{0.760} & \scalebox{0.8}{0.473} &{\scalebox{0.8}{0.620}} &{\scalebox{0.8}{0.336}} &\scalebox{0.8}{0.625} &\scalebox{0.8}{0.383} & \scalebox{0.8}{0.804} & \scalebox{0.8}{0.509} &{\scalebox{0.8}{0.610}} &\scalebox{0.8}{0.376} &\scalebox{0.8}{0.624} &{\scalebox{0.8}{0.340}} &\scalebox{0.8}{0.628} &\scalebox{0.8}{0.379} \\
    
    \midrule
    \scalebox{0.95}{Weather} & \boldres{\scalebox{0.8}{0.242}} & \boldres{\scalebox{0.8}{0.271}} & \secondres{\scalebox{0.8}{0.258}} & \secondres{\scalebox{0.8}{0.278}} &\scalebox{0.8}{0.272} &\scalebox{0.8}{0.291} & {\scalebox{0.8}{0.259}} & {\scalebox{0.8}{0.281}} & \scalebox{0.8}{0.259} & \scalebox{0.8}{0.315}  & \scalebox{0.8}{0.271} & \scalebox{0.8}{0.320} &{\scalebox{0.8}{0.259}} &{\scalebox{0.8}{0.287}} &\scalebox{0.8}{0.265} &\scalebox{0.8}{0.317} & \scalebox{0.8}{0.292} & \scalebox{0.8}{0.363} &\scalebox{0.8}{0.309} &\scalebox{0.8}{0.360} &\scalebox{0.8}{0.288} &\scalebox{0.8}{0.314} &\scalebox{0.8}{0.338} &\scalebox{0.8}{0.382} \\
    \midrule
    \scalebox{0.95}{Solar-Energy} & \boldres{\scalebox{0.8}{0.225}} & \boldres{\scalebox{0.8}{0.260}} &\secondres{\scalebox{0.8}{0.233}} &\secondres{\scalebox{0.8}{0.262}} &\scalebox{0.8}{0.369} &\scalebox{0.8}{0.356} &{\scalebox{0.8}{0.270}} & {\scalebox{0.8}{0.307}} &\scalebox{0.8}{0.641} &\scalebox{0.8}{0.639} &\scalebox{0.8}{0.347} &\scalebox{0.8}{0.417} &\scalebox{0.8}{0.301} &\scalebox{0.8}{0.319} &\scalebox{0.8}{0.330} &\scalebox{0.8}{0.401} &\scalebox{0.8}{0.282} &\scalebox{0.8}{0.375} &\scalebox{0.8}{0.291} &\scalebox{0.8}{0.381} &\scalebox{0.8}{0.261} &\scalebox{0.8}{0.381} &\scalebox{0.8}{0.885} &\scalebox{0.8}{0.711} \\
    \midrule
    \midrule
    \scalebox{0.95}{{$1^{\text{st}}$ Count}} & \boldres{\scalebox{0.8}{7}} & \boldres{\scalebox{0.8}{8}} &\secondres{\scalebox{0.8}{1}} & \scalebox{0.8}{0} &\scalebox{0.8}{0} & \secondres{\scalebox{0.8}{1}} &\scalebox{0.8}{0} & \secondres{\scalebox{0.8}{1}} &\scalebox{0.8}{0} &\scalebox{0.8}{0} & \scalebox{0.8}{0}& \scalebox{0.8}{0} &\scalebox{0.8}{0} &\scalebox{0.8}{0} &\scalebox{0.8}{0} &\scalebox{0.8}{0} &\scalebox{0.8}{0}&\scalebox{0.8}{0} &\secondres{\scalebox{0.8}{1}} &\scalebox{0.8}{0} &\scalebox{0.8}{0} &\scalebox{0.8}{0} &\scalebox{0.8}{0} &\scalebox{0.8}{0} \\
    \bottomrule
  \end{tabular}
    \end{small}
  \end{threeparttable}
}
\end{table}
\paragraph{Long-term forecasting}
Following iTransformer \citep{iTransformer}, we use 4 different prediction lengths (i.e., \{96, 192, 336, 720\}) and fix the lookback window length as 96 for the long-term forecasting task. We evaluate models with MSE (Mean Squared Error) and MAE (Mean Absolute Error) -- the lower values indicate better prediction performance. 
We summarize the long-term forecasting results in Table~\ref{tab:main_result_long_term} with the best in \boldres{red} and the second \secondres{underlined}.
Overall, we can see that \ourmodel achieves the best results compared with 11 baselines on 7 out of 9 datasets for MSE and 8 out of 9 datasets for MAE. 
Particularly, iTransformer, as the previous state-of-the-art model, performs worse than our model in most cases of ETT datasets and ECL dataset (which are both from electricity domain). This may indicate that only model multivariate correlation without considering temporal correlation is not effective for some datasets.
Meanwhile, the results of PatchTST are also deficient, suggesting that only capturing temporal relationships within a channel is not sufficient as well.
In contrast, our proposed model \ourmodel can better capture temporal relationships both within a variate and across different variates, which leads to better prediction performance.  
% both inter-channel and intra-channel. 
% \TODO{To discuss about Crossformer here as well, why it performs bad. }
Besides, although Crossformer is claimed to capture cross-time and cross-variate dependencies, it still performs much worse compared with our approach.
The reason is that their sequential design with two attention modules cannot simultaneously and effectively capture cross-time and cross-variate dependencies, while our approach can explicitly model these dependencies at the same time. 

\paragraph{Short-term forecasting}
Besides long-term forecasting, we also conduct experiments for short-term forecasting with 4 prediction lengths (i.e., \{12, 24, 48, 96\}) on PEMS datasets as in SCINet~\citep{SCINet} and iTransformer~\citep{iTransformer}. 
Full results on 4 PEMS datasets with 4 different prediction lengths are shown in Table~\ref{tab:full_baseline_results_pems}. 
Generally, our model outperforms other baselines on all prediction lengths and all PEMS datasets, which demonstrates the superiority of capturing cross-channel cross-time relationships for short-term forecasting. 
Additionally, we observe that PatchTST usually underperforms iTransformer by a large margin, suggesting that modeling channel dependencies is necessary for PEMS datasets. 
The worse results of iTransformer, compared with our model, indicate that cross-channel temporal relationships are important and should be captured on these datasets.

\begin{table}[htbp]
  \caption{\update{Full results of the PEMS forecasting task}. We compare extensive competitive models under different prediction lengths following the setting of SCINet~\citeyearpar{SCINet}. The input length is set to 96 for all baselines. \emph{Avg} means the average results from all four prediction lengths.
  }
  \label{tab:full_baseline_results_pems}
  \vskip -0.0in
  \vspace{3pt}
  \renewcommand{\arraystretch}{0.85} 
  \centering
  \resizebox{1\columnwidth}{!}{
  \begin{threeparttable}
  \begin{small}
  \renewcommand{\multirowsetup}{\centering}
  \setlength{\tabcolsep}{1pt}
  \begin{tabular}{c|c|cc|cc|cc|cc|cc|cc|cc|cc|cc|cc|cc|cc}
    \toprule
    \multicolumn{2}{c}{\multirow{2}{*}{Models}} & 
    \multicolumn{2}{c}{\rotatebox{0}{\scalebox{0.8}{\textbf{\ourmodel}}}} &
    \multicolumn{2}{c}{\rotatebox{0}{\scalebox{0.8}{iTransformer}}} &
    \multicolumn{2}{c}{\rotatebox{0}{\scalebox{0.8}{\update{RLinear}}}} &
    \multicolumn{2}{c}{\rotatebox{0}{\scalebox{0.8}{PatchTST}}} &
    \multicolumn{2}{c}{\rotatebox{0}{\scalebox{0.8}{Crossformer}}} &
    \multicolumn{2}{c}{\rotatebox{0}{\scalebox{0.8}{TiDE}}} &
    \multicolumn{2}{c}{\rotatebox{0}{\scalebox{0.8}{{TimesNet}}}} &
    \multicolumn{2}{c}{\rotatebox{0}{\scalebox{0.8}{DLinear}}} &
    \multicolumn{2}{c}{\rotatebox{0}{\scalebox{0.8}{SCINet}}} &
    \multicolumn{2}{c}{\rotatebox{0}{\scalebox{0.8}{FEDformer}}} &
    \multicolumn{2}{c}{\rotatebox{0}{\scalebox{0.8}{Stationary}}} &
    \multicolumn{2}{c}{\rotatebox{0}{\scalebox{0.8}{Autoformer}}} \\% &\multicolumn{2}{c}{\rotatebox{0}{\scalebox{0.8}{Informer}}} \\
    \multicolumn{2}{c}{} &
    \multicolumn{2}{c}{\scalebox{0.8}{\textbf{(Ours)}}} & 
    \multicolumn{2}{c}{\scalebox{0.8}{\citeyearpar{li2023revisiting}}} &
    \multicolumn{2}{c}{\scalebox{0.8}{\citeyearpar{li2023revisiting}}} & 
    \multicolumn{2}{c}{\scalebox{0.8}{\citeyearpar{PatchTST}}} & 
    \multicolumn{2}{c}{\scalebox{0.8}{\citeyearpar{Crossformer}}} & 
    \multicolumn{2}{c}{\scalebox{0.8}{\citeyearpar{das2023long}}} & 
    \multicolumn{2}{c}{\scalebox{0.8}{\citeyearpar{Timesnet}}} & 
    \multicolumn{2}{c}{\scalebox{0.8}{\citeyearpar{DLinear}}} &
    \multicolumn{2}{c}{\scalebox{0.8}{\citeyearpar{SCINet}}} & 
    \multicolumn{2}{c}{\scalebox{0.8}{\citeyearpar{fedformer}}} &
    \multicolumn{2}{c}{\scalebox{0.8}{\citeyearpar{Stationary}}} &
    \multicolumn{2}{c}{\scalebox{0.8}{\citeyearpar{Autoformer}}} 
    \\         %& \multicolumn{2}{c}{\scalebox{0.8}{\citeyearpar{Informer}}} \\
    \cmidrule(lr){3-4} \cmidrule(lr){5-6}\cmidrule(lr){7-8} \cmidrule(lr){9-10}\cmidrule(lr){11-12}\cmidrule(lr){13-14} \cmidrule(lr){15-16} \cmidrule(lr){17-18} \cmidrule(lr){19-20} \cmidrule(lr){21-22} \cmidrule(lr){23-24} \cmidrule(lr){25-26}
    \multicolumn{2}{c}{Metric}  & \scalebox{0.78}{MSE} & \scalebox{0.78}{MAE}  & \scalebox{0.78}{MSE} & \scalebox{0.78}{MAE}  & \scalebox{0.78}{MSE} & \scalebox{0.78}{MAE}  & \scalebox{0.78}{MSE} & \scalebox{0.78}{MAE}  & \scalebox{0.78}{MSE} & \scalebox{0.78}{MAE}  & \scalebox{0.78}{MSE} & \scalebox{0.78}{MAE} & \scalebox{0.78}{MSE} & \scalebox{0.78}{MAE} & \scalebox{0.78}{MSE} & \scalebox{0.78}{MAE} & \scalebox{0.78}{MSE} & \scalebox{0.78}{MAE} & \scalebox{0.78}{MSE} & \scalebox{0.78}{MAE} & \scalebox{0.78}{MSE} & \scalebox{0.78}{MAE} & \scalebox{0.78}{MSE} & \scalebox{0.78}{MAE}\\
    \toprule
    
    \multirow{5}{*}{\rotatebox{90}{\scalebox{0.95}{PEMS03}}}
    &  \scalebox{0.78}{12} & \boldres{\scalebox{0.78}{0.059} }& \boldres{\scalebox{0.78}{0.160} }&{\scalebox{0.78}{0.071}} &{\scalebox{0.78}{0.174}} &\scalebox{0.78}{0.126} &\scalebox{0.78}{0.236} &\scalebox{0.78}{0.099} &\scalebox{0.78}{0.216} &\scalebox{0.78}{0.090} &\scalebox{0.78}{0.203} & \scalebox{0.78}{0.178} & \scalebox{0.78}{0.305} &\scalebox{0.78}{0.085} &\scalebox{0.78}{0.192} &\scalebox{0.78}{0.122} &\scalebox{0.78}{0.243} &\secondres{\scalebox{0.78}{0.066}} &\secondres{\scalebox{0.78}{0.172}} &\scalebox{0.78}{0.126} &\scalebox{0.78}{0.251} &\scalebox{0.78}{0.081} &\scalebox{0.78}{0.188} &\scalebox{0.78}{0.272} &\scalebox{0.78}{0.385}    \\% &\scalebox{0.78}{0.126} &\scalebox{0.78}{0.233}\\
    & \scalebox{0.78}{24} &\boldres{\scalebox{0.78}{0.074}}& \boldres{\scalebox{0.78}{0.180}}&\scalebox{0.78}{0.093} &\scalebox{0.78}{0.201} &\scalebox{0.78}{0.246} &\scalebox{0.78}{0.334} &\scalebox{0.78}{0.142} &\scalebox{0.78}{0.259} &\scalebox{0.78}{0.121} &\scalebox{0.78}{0.240} & \scalebox{0.78}{0.257} & \scalebox{0.78}{0.371} &\scalebox{0.78}{0.118} &\scalebox{0.78}{0.223} &\scalebox{0.78}{0.201} &\scalebox{0.78}{0.317} &\secondres{\scalebox{0.78}{0.085}} &\secondres{\scalebox{0.78}{0.198}} &\scalebox{0.78}{0.149} &\scalebox{0.78}{0.275} &\scalebox{0.78}{0.105} &\scalebox{0.78}{0.214} &\scalebox{0.78}{0.334} &\scalebox{0.78}{0.440} \\%&\scalebox{0.78}{0.139} &\scalebox{0.78}{0.250}\\
    & \scalebox{0.78}{48} & \boldres{\scalebox{0.78}{0.104} }& \boldres{\scalebox{0.78}{0.213} }& \secondres{\scalebox{0.78}{0.125}} &\secondres{\scalebox{0.78}{0.236}} &\scalebox{0.78}{0.551} &\scalebox{0.78}{0.529} &\scalebox{0.78}{0.211} &\scalebox{0.78}{0.319}  &\scalebox{0.78}{0.202} &\scalebox{0.78}{0.317} & \scalebox{0.78}{0.379}& \scalebox{0.78}{0.463} &\scalebox{0.78}{0.155} &\scalebox{0.78}{0.260} &\scalebox{0.78}{0.333} &\scalebox{0.78}{0.425} &{\scalebox{0.78}{0.127}} &{\scalebox{0.78}{0.238}} &\scalebox{0.78}{0.227} &\scalebox{0.78}{0.348} &\scalebox{0.78}{0.154} &\scalebox{0.78}{0.257} &\scalebox{0.78}{1.032} &\scalebox{0.78}{0.782} \\% &\scalebox{0.78}{0.186} &\scalebox{0.78}{0.289}\\
    & \scalebox{0.78}{96} & \boldres{\scalebox{0.78}{0.151} }& \boldres{\scalebox{0.78}{0.261} }&\secondres{\scalebox{0.78}{0.164}} &\secondres{\scalebox{0.78}{0.275}} &\scalebox{0.78}{1.057} &\scalebox{0.78}{0.787} &\scalebox{0.78}{0.269} &\scalebox{0.78}{0.370} &\scalebox{0.78}{0.262} &\scalebox{0.78}{0.367} & \scalebox{0.78}{0.490}& \scalebox{0.78}{0.539} &\scalebox{0.78}{0.228} &\scalebox{0.78}{0.317} &\scalebox{0.78}{0.457} &\scalebox{0.78}{0.515} &{\scalebox{0.78}{0.178}} &{\scalebox{0.78}{0.287}} &\scalebox{0.78}{0.348} &\scalebox{0.78}{0.434} &\scalebox{0.78}{0.247} &\scalebox{0.78}{0.336} &\scalebox{0.78}{1.031} &\scalebox{0.78}{0.796} \\% &\scalebox{0.78}{0.233} &\scalebox{0.78}{0.323}\\
    \cmidrule(lr){2-26}
    & \scalebox{0.78}{Avg} & \boldres{\scalebox{0.78}{0.097} }& \boldres{\scalebox{0.78}{0.204} }&\secondres{\scalebox{0.78}{0.113}} &\secondres{\scalebox{0.78}{0.221}} &\scalebox{0.78}{0.495} &\scalebox{0.78}{0.472} &\scalebox{0.78}{0.180} &\scalebox{0.78}{0.291} &\scalebox{0.78}{0.169} &\scalebox{0.78}{0.281} & \scalebox{0.78}{0.326}& \scalebox{0.78}{0.419} &\scalebox{0.78}{0.147} &\scalebox{0.78}{0.248} &\scalebox{0.78}{0.278} &\scalebox{0.78}{0.375} &{\scalebox{0.78}{0.114}} &{\scalebox{0.78}{0.224}} &\scalebox{0.78}{0.213} &\scalebox{0.78}{0.327} &\scalebox{0.78}{0.147} &\scalebox{0.78}{0.249} &\scalebox{0.78}{0.667} &\scalebox{0.78}{0.601} \\% &\scalebox{0.78}{0.171} &\scalebox{0.78}{0.274}\\
    
    \midrule
    \multirow{5}{*}{\update{\rotatebox{90}{\scalebox{0.95}{PEMS04}}}} 
    &  \scalebox{0.78}{12}  & \boldres{\scalebox{0.78}{0.070} }& \boldres{\scalebox{0.78}{0.172}  }&{\scalebox{0.78}{0.078}} &{\scalebox{0.78}{0.183}} &\scalebox{0.78}{0.138} &\scalebox{0.78}{0.252} &\scalebox{0.78}{0.105} &\scalebox{0.78}{0.224} &\scalebox{0.78}{0.098} &\scalebox{0.78}{0.218} & \scalebox{0.78}{0.219}& \scalebox{0.78}{0.340} &\scalebox{0.78}{0.087} &\scalebox{0.78}{0.195} &\scalebox{0.78}{0.148} &\scalebox{0.78}{0.272} &\secondres{\scalebox{0.78}{0.073}} &\secondres{\scalebox{0.78}{0.177}} &\scalebox{0.78}{0.138} &\scalebox{0.78}{0.262} &\scalebox{0.78}{0.088} &\scalebox{0.78}{0.196} &\scalebox{0.78}{0.424} &\scalebox{0.78}{0.491} \\% &\scalebox{0.78}{0.112} &\scalebox{0.78}{0.222}\\
    & \scalebox{0.78}{24}  & \boldres{\scalebox{0.78}{0.082} }& \boldres{\scalebox{0.78}{0.189}  }&{\scalebox{0.78}{0.095}} &{\scalebox{0.78}{0.205}} &\scalebox{0.78}{0.258} &\scalebox{0.78}{0.348} &\scalebox{0.78}{0.153} &\scalebox{0.78}{0.275} &\scalebox{0.78}{0.131} &\scalebox{0.78}{0.256} & \scalebox{0.78}{0.292}& \scalebox{0.78}{0.398} &\scalebox{0.78}{0.103} &\scalebox{0.78}{0.215} &\scalebox{0.78}{0.224} &\scalebox{0.78}{0.340} &\secondres{\scalebox{0.78}{0.084}} &\secondres{\scalebox{0.78}{0.193}} &\scalebox{0.78}{0.177} &\scalebox{0.78}{0.293} &\scalebox{0.78}{0.104} &\scalebox{0.78}{0.216} &\scalebox{0.78}{0.459} &\scalebox{0.78}{0.509} \\% &\scalebox{0.78}{0.117} &\scalebox{0.78}{0.227}\\
    & \scalebox{0.78}{48}  & \secondres{\scalebox{0.78}{0.104} }& \secondres{\scalebox{0.78}{0.216}  }&{\scalebox{0.78}{0.120}} &{\scalebox{0.78}{0.233}} &\scalebox{0.78}{0.572} &\scalebox{0.78}{0.544} &\scalebox{0.78}{0.229} &\scalebox{0.78}{0.339} &\scalebox{0.78}{0.205} &\scalebox{0.78}{0.326} & \scalebox{0.78}{0.409}& \scalebox{0.78}{0.478} &\scalebox{0.78}{0.136} &\scalebox{0.78}{0.250} &\scalebox{0.78}{0.355} &\scalebox{0.78}{0.437} &\boldres{\scalebox{0.78}{0.099}} &\boldres{\scalebox{0.78}{0.211}} &\scalebox{0.78}{0.270} &\scalebox{0.78}{0.368} &\scalebox{0.78}{0.137} &\scalebox{0.78}{0.251} &\scalebox{0.78}{0.646} &\scalebox{0.78}{0.610} \\% &\scalebox{0.78}{0.126} &\scalebox{0.78}{0.239}\\
    & \scalebox{0.78}{96}  & \secondres{\scalebox{0.78}{0.137} }& \secondres{\scalebox{0.78}{0.256}  }&{\scalebox{0.78}{0.150}} &{\scalebox{0.78}{0.262}} &\scalebox{0.78}{1.137} &\scalebox{0.78}{0.820} &\scalebox{0.78}{0.291} &\scalebox{0.78}{0.389} &\scalebox{0.78}{0.402} &\scalebox{0.78}{0.457} & \scalebox{0.78}{0.492}& \scalebox{0.78}{0.532} &\scalebox{0.78}{0.190} &\scalebox{0.78}{0.303} &\scalebox{0.78}{0.452} &\scalebox{0.78}{0.504} &\boldres{\scalebox{0.78}{0.114}} &\boldres{\scalebox{0.78}{0.227}} &\scalebox{0.78}{0.341} &\scalebox{0.78}{0.427} &\scalebox{0.78}{0.186} &\scalebox{0.78}{0.297} &\scalebox{0.78}{0.912} &\scalebox{0.78}{0.748} \\% &\scalebox{0.78}{0.128} &\scalebox{0.78}{0.242}\\
    \cmidrule(lr){2-26}
    & \scalebox{0.78}{Avg} & \secondres{\scalebox{0.78}{0.098} }& \secondres{\scalebox{0.78}{0.208} }&{\scalebox{0.78}{0.111}} &{\scalebox{0.78}{0.221}} &\scalebox{0.78}{0.526} &\scalebox{0.78}{0.491} &\scalebox{0.78}{0.195} &\scalebox{0.78}{0.307} &\scalebox{0.78}{0.209} &\scalebox{0.78}{0.314} & \scalebox{0.78}{0.353}& \scalebox{0.78}{0.437} &\scalebox{0.78}{0.129} &\scalebox{0.78}{0.241} &\scalebox{0.78}{0.295} &\scalebox{0.78}{0.388} &\boldres{\scalebox{0.78}{0.092}} &\boldres{\scalebox{0.78}{0.202}} &\scalebox{0.78}{0.231} &\scalebox{0.78}{0.337} &\scalebox{0.78}{0.127} &\scalebox{0.78}{0.240} &\scalebox{0.78}{0.610} &\scalebox{0.78}{0.590} \\% &\scalebox{0.78}{0.121} &\scalebox{0.78}{0.232}\\

    \midrule
    \multirow{5}{*}{\update{\rotatebox{90}{\scalebox{0.95}{PEMS07}}}}
    &  \scalebox{0.78}{12} & \boldres{\scalebox{0.78}{0.057} }& \boldres{\scalebox{0.78}{0.153} }&\secondres{\scalebox{0.78}{0.067}} &\secondres{\scalebox{0.78}{0.165}} &\scalebox{0.78}{0.118} &\scalebox{0.78}{0.235} &\scalebox{0.78}{0.095} &\scalebox{0.78}{0.207} &\scalebox{0.78}{0.094} &\scalebox{0.78}{0.200} & \scalebox{0.78}{0.173}& \scalebox{0.78}{0.304} &\scalebox{0.78}{0.082} &\scalebox{0.78}{0.181} &\scalebox{0.78}{0.115} &\scalebox{0.78}{0.242} &{\scalebox{0.78}{0.068}} &{\scalebox{0.78}{0.171}} &\scalebox{0.78}{0.109} &\scalebox{0.78}{0.225} &\scalebox{0.78}{0.083} &\scalebox{0.78}{0.185} &\scalebox{0.78}{0.199} &\scalebox{0.78}{0.336} \\% &\scalebox{0.78}{0.173} &\scalebox{0.78}{0.243} \\
    & \scalebox{0.78}{24} & \boldres{\scalebox{0.78}{0.075} }& \boldres{\scalebox{0.78}{0.174} }&\secondres{\scalebox{0.78}{0.088}} &\secondres{\scalebox{0.78}{0.190}} &\scalebox{0.78}{0.242} &\scalebox{0.78}{0.341} &\scalebox{0.78}{0.150} &\scalebox{0.78}{0.262} &\scalebox{0.78}{0.139} &\scalebox{0.78}{0.247} & \scalebox{0.78}{0.271}& \scalebox{0.78}{0.383} &\scalebox{0.78}{0.101} &\scalebox{0.78}{0.204} &\scalebox{0.78}{0.210} &\scalebox{0.78}{0.329} &{\scalebox{0.78}{0.119}} &{\scalebox{0.78}{0.225}} &\scalebox{0.78}{0.125} &\scalebox{0.78}{0.244} &\scalebox{0.78}{0.102} &\scalebox{0.78}{0.207} &\scalebox{0.78}{0.323} &\scalebox{0.78}{0.420} \\% &\scalebox{0.78}{0.178} &\scalebox{0.78}{0.247}\\
    & \scalebox{0.78}{48} & \boldres{\scalebox{0.78}{0.107} }& \boldres{\scalebox{0.78}{0.208} }&\secondres{\scalebox{0.78}{0.110}} &\secondres{\scalebox{0.78}{0.215}} &\scalebox{0.78}{0.562} &\scalebox{0.78}{0.541} &\scalebox{0.78}{0.253} &\scalebox{0.78}{0.340} &\scalebox{0.78}{0.311} &\scalebox{0.78}{0.369} & \scalebox{0.78}{0.446}& \scalebox{0.78}{0.495} &\scalebox{0.78}{0.134} &\scalebox{0.78}{0.238} &\scalebox{0.78}{0.398} &\scalebox{0.78}{0.458} &{\scalebox{0.78}{0.149}} &{\scalebox{0.78}{0.237}} &\scalebox{0.78}{0.165} &\scalebox{0.78}{0.288} &\scalebox{0.78}{0.136} &\scalebox{0.78}{0.240} &\scalebox{0.78}{0.390} &\scalebox{0.78}{0.470} \\% &\scalebox{0.78}{0.185} &\scalebox{0.78}{0.256}\\
    & \scalebox{0.78}{96} & \boldres{\scalebox{0.78}{0.133} }& \boldres{\scalebox{0.78}{0.228} }&\secondres{\scalebox{0.78}{0.139}} &\secondres{\scalebox{0.78}{0.245}} &\scalebox{0.78}{1.096} &\scalebox{0.78}{0.795} &\scalebox{0.78}{0.346} &\scalebox{0.78}{0.404} &\scalebox{0.78}{0.396} &\scalebox{0.78}{0.442} & \scalebox{0.78}{0.628}& \scalebox{0.78}{0.577} &\scalebox{0.78}{0.181} &\scalebox{0.78}{0.279} &\scalebox{0.78}{0.594} &\scalebox{0.78}{0.553} &{\scalebox{0.78}{0.141}} &{\scalebox{0.78}{0.234}} &\scalebox{0.78}{0.262} &\scalebox{0.78}{0.376} &\scalebox{0.78}{0.187} &\scalebox{0.78}{0.287} &\scalebox{0.78}{0.554} &\scalebox{0.78}{0.578} \\% &\scalebox{0.78}{0.195} &\scalebox{0.78}{0.269}\\
    \cmidrule(lr){2-26}
    & \scalebox{0.78}{Avg} & \boldres{\scalebox{0.78}{0.093} }& \boldres{\scalebox{0.78}{0.191} }&\secondres{\scalebox{0.78}{0.101}} &\secondres{\scalebox{0.78}{0.204}} &\scalebox{0.78}{0.504} &\scalebox{0.78}{0.478} &\scalebox{0.78}{0.211} &\scalebox{0.78}{0.303} &\scalebox{0.78}{0.235} &\scalebox{0.78}{0.315} & \scalebox{0.78}{0.380}& \scalebox{0.78}{0.440} &\scalebox{0.78}{0.124} &\scalebox{0.78}{0.225} &\scalebox{0.78}{0.329} &\scalebox{0.78}{0.395} &{\scalebox{0.78}{0.119}} &{\scalebox{0.78}{0.234}} &\scalebox{0.78}{0.165} &\scalebox{0.78}{0.283} &\scalebox{0.78}{0.127} &\scalebox{0.78}{0.230} &\scalebox{0.78}{0.367} &\scalebox{0.78}{0.451} \\% &\scalebox{0.78}{0.183} &\scalebox{0.78}{0.254}\\

    \midrule
    \multirow{5}{*}{\update{\rotatebox{90}{\scalebox{0.95}{PEMS08}}}}
    &  \scalebox{0.78}{12}& \boldres{\scalebox{0.78}{0.073} }& \boldres{\scalebox{0.78}{0.174}  }&\secondres{\scalebox{0.78}{0.079}} &\secondres{\scalebox{0.78}{0.182}} &\scalebox{0.78}{0.133} &\scalebox{0.78}{0.247} &\scalebox{0.78}{0.168} &\scalebox{0.78}{0.232} &\scalebox{0.78}{0.165} &\scalebox{0.78}{0.214} & \scalebox{0.78}{0.227}& \scalebox{0.78}{0.343} &\scalebox{0.78}{0.112} &\scalebox{0.78}{0.212} &\scalebox{0.78}{0.154} &\scalebox{0.78}{0.276} &{\scalebox{0.78}{0.087}} &{\scalebox{0.78}{0.184}} &\scalebox{0.78}{0.173} &\scalebox{0.78}{0.273} &\scalebox{0.78}{0.109} &\scalebox{0.78}{0.207} &\scalebox{0.78}{0.436} &\scalebox{0.78}{0.485} \\% &\scalebox{0.78}{0.296} &\scalebox{0.78}{0.312}\\
    & \scalebox{0.78}{24} & \boldres{\scalebox{0.78}{0.096} }& \boldres{\scalebox{0.78}{0.197} }&\secondres{\scalebox{0.78}{0.115}} &\secondres{\scalebox{0.78}{0.219}} &\scalebox{0.78}{0.249} &\scalebox{0.78}{0.343} &\scalebox{0.78}{0.224} &\scalebox{0.78}{0.281} &\scalebox{0.78}{0.215} &\scalebox{0.78}{0.260} & \scalebox{0.78}{0.318}& \scalebox{0.78}{0.409} &\scalebox{0.78}{0.141} &\scalebox{0.78}{0.238} &\scalebox{0.78}{0.248} &\scalebox{0.78}{0.353} &{\scalebox{0.78}{0.122}} &{\scalebox{0.78}{0.221}} &\scalebox{0.78}{0.210} &\scalebox{0.78}{0.301} &\scalebox{0.78}{0.140} &\scalebox{0.78}{0.236} &\scalebox{0.78}{0.467} &\scalebox{0.78}{0.502} \\% &\scalebox{0.78}{0.327} &\scalebox{0.78}{0.318}\\
    & \scalebox{0.78}{48} & \boldres{\scalebox{0.78}{0.141} }& \boldres{\scalebox{0.78}{0.239} }&\secondres{\scalebox{0.78}{0.186}} &\secondres{\scalebox{0.78}{0.235}} &\scalebox{0.78}{0.569} &\scalebox{0.78}{0.544} &\scalebox{0.78}{0.321} &\scalebox{0.78}{0.354} &\scalebox{0.78}{0.315} &\scalebox{0.78}{0.355} & \scalebox{0.78}{0.497}& \scalebox{0.78}{0.510} &\scalebox{0.78}{0.198} &\scalebox{0.78}{0.283} &\scalebox{0.78}{0.440} &\scalebox{0.78}{0.470} &{\scalebox{0.78}{0.189}} &{\scalebox{0.78}{0.270}} &\scalebox{0.78}{0.320} &\scalebox{0.78}{0.394} &\scalebox{0.78}{0.211} &\scalebox{0.78}{0.294} &\scalebox{0.78}{0.966} &\scalebox{0.78}{0.733} \\% &\scalebox{0.78}{0.387} &\scalebox{0.78}{0.365}\\
    & \scalebox{0.78}{96} & \boldres{\scalebox{0.78}{0.210} }& \boldres{\scalebox{0.78}{0.275} }&\secondres{\scalebox{0.78}{0.221}} &\secondres{\scalebox{0.78}{0.267}} &\scalebox{0.78}{1.166} &\scalebox{0.78}{0.814} &\scalebox{0.78}{0.408} &\scalebox{0.78}{0.417} &\scalebox{0.78}{0.377} &\scalebox{0.78}{0.397} & \scalebox{0.78}{0.721}& \scalebox{0.78}{0.592} &\scalebox{0.78}{0.320} &\scalebox{0.78}{0.351} &\scalebox{0.78}{0.674} &\scalebox{0.78}{0.565} &{\scalebox{0.78}{0.236}} &{\scalebox{0.78}{0.300}} &\scalebox{0.78}{0.442} &\scalebox{0.78}{0.465} &\scalebox{0.78}{0.345} &\scalebox{0.78}{0.367} &\scalebox{0.78}{1.385} &\scalebox{0.78}{0.915} \\% &\scalebox{0.78}{0.455} &\scalebox{0.78}{0.407}\\
    \cmidrule(lr){2-26}
    & \scalebox{0.78}{Avg} & \boldres{\scalebox{0.78}{0.130} }& \boldres{\scalebox{0.78}{0.221} }&\secondres{\scalebox{0.78}{0.150}} &\secondres{\scalebox{0.78}{0.226}} &\scalebox{0.78}{0.529} &\scalebox{0.78}{0.487} &\scalebox{0.78}{0.280} &\scalebox{0.78}{0.321} &\scalebox{0.78}{0.268} &\scalebox{0.78}{0.307} & \scalebox{0.78}{0.441}& \scalebox{0.78}{0.464} &\scalebox{0.78}{0.193} &\scalebox{0.78}{0.271} &\scalebox{0.78}{0.379} &\scalebox{0.78}{0.416} &{\scalebox{0.78}{0.158}} &{\scalebox{0.78}{0.244}} &\scalebox{0.78}{0.286} &\scalebox{0.78}{0.358} &\scalebox{0.78}{0.201} &\scalebox{0.78}{0.276} &\scalebox{0.78}{0.814} &\scalebox{0.78}{0.659} \\% &\scalebox{0.78}{0.366} &\scalebox{0.78}{0.350}\\
    \midrule
     \multicolumn{2}{c|}{\scalebox{0.78}{{$1^{\text{st}}$ Count}}} & \boldres{\scalebox{0.78}{14} }& \boldres{\scalebox{0.78}{14} }& \scalebox{0.78}{0} & \scalebox{0.78}{0} & \scalebox{0.78}{0}& \scalebox{0.78}{0}& \scalebox{0.78}{0}& \scalebox{0.78}{0}& \scalebox{0.78}{0}& \scalebox{0.78}{0}& \scalebox{0.78}{0}& \scalebox{0.78}{0}& \scalebox{0.78}{0}& \scalebox{0.78}{0}& \scalebox{0.78}{0}& \scalebox{0.78}{0}& \scalebox{0.78}{\secondres{2}} & \scalebox{0.78}{\secondres{2}} & \scalebox{0.78}{0}& \scalebox{0.78}{0}& \scalebox{0.78}{0}& \scalebox{0.78}{0}& \scalebox{0.78}{0}& \scalebox{0.78}{0} \\% & \scalebox{0.78}{0}& \scalebox{0.78}{0}\\

        \bottomrule
      \end{tabular}
    \end{small}
  \end{threeparttable}
}
\end{table}

% 1) main experiments to the good performance of our model (DONE) (separate the results into two types 1. long-term forecasting, 2. short-term forecasting PEMS datasets, as in TimeMixer paper).

\subsection{Model Analysis}
\paragraph{Ablation study}
% \usepackage{booktabs}
\begin{table}[]
\caption{The effectiveness of our dispatcher module. OOM indicates the ``Out of Memory'' error on GPUs (we a single A100 GPU of memory 40GB).}
\label{tab: ablation_dispatchers}
\centering
\begin{tabular}{c|cc|cc|cc|cc}
\toprule
                & \multicolumn{2}{c|}{ETTm1} & \multicolumn{2}{c|}{Weather} & \multicolumn{2}{c|}{ECL} & \multicolumn{2}{c}{Traffic} \\
                & MSE          & Mem         & MSE           & Mem          & MSE         & Mem        & MSE          & Mem          \\ \midrule
w/o dispatchers & 0.385        & 2.56GB        & 0.247         & 9.17GB         & OOM         & OOM        & OOM          & OOM          \\
w/ dispatchers  & 0.379        & 2.33GB        & 0.242         & 5.13GB         & 0.166       & 13.32GB      & 0.439        & 22.87GB        \\ \bottomrule
\end{tabular}
\end{table}
% \TODO{Put a table to compare the performance of our model with dispatcher and without dispatcher. Meanwhile, provide the peak GPU memory usage to show that we have reduced the memory usage.}
We conduct the ablation study to verify the effectiveness of our dispatcher module by using the same setting (e.g., the number of layers, hidden dimensions, batch size) for comparing the our model with and without dispatchers. 
In Table \ref{tab: ablation_dispatchers}, we can see that adding dispatchers helps to reduce GPU usage. In ECL and Traffic, the version without dispatchers even leads to out-of-memory (OOM) issues. 
Moreover, we observe that the memory reduction becomes more significant when the number of variates increases. On ETTm1 with 7 variates, the memory only reduces from 2.56GB to 2.33GB, while on ECL and Traffic, it reduces from OOM (more than 40GB) to 13.32GB and 22.87GB, respectively.

% 2) \TODO{synthetic experiments to directly support our claim (i.e., our model has the ability to capture diverse channel and temporal dependencies)... The current results are not very ideal...} 

% 3) Side experiments / further investigation. 

\begin{figure}[ht]
    \centering
    \begin{subfigure}{0.45\textwidth}
        \centering
        \includegraphics[width=\linewidth]{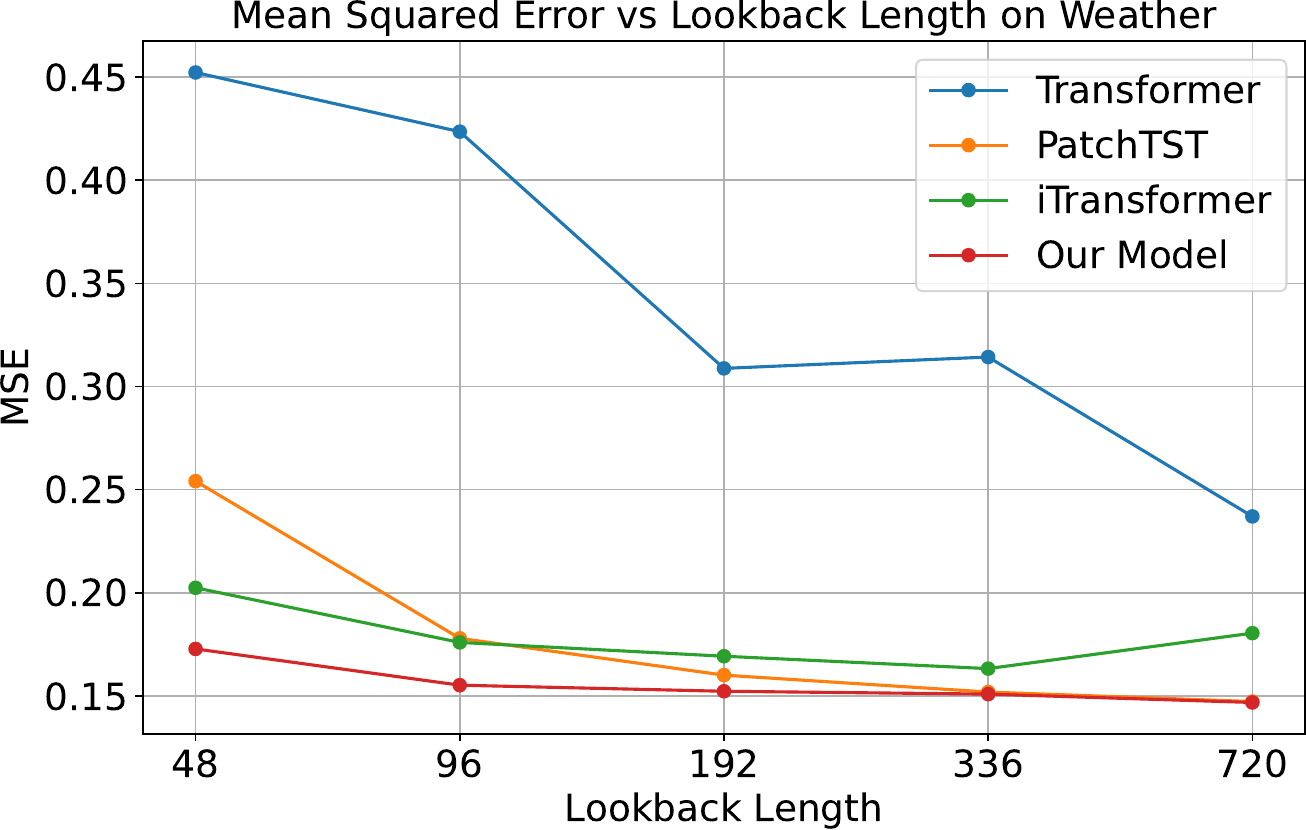}
        % \caption{TBD (\TODO{Change the font size of plots)}}
        % \label{fig:sub1}
    \end{subfigure}
    \hfill
    \begin{subfigure}{0.45\textwidth}
        \centering
        \includegraphics[width=\linewidth]{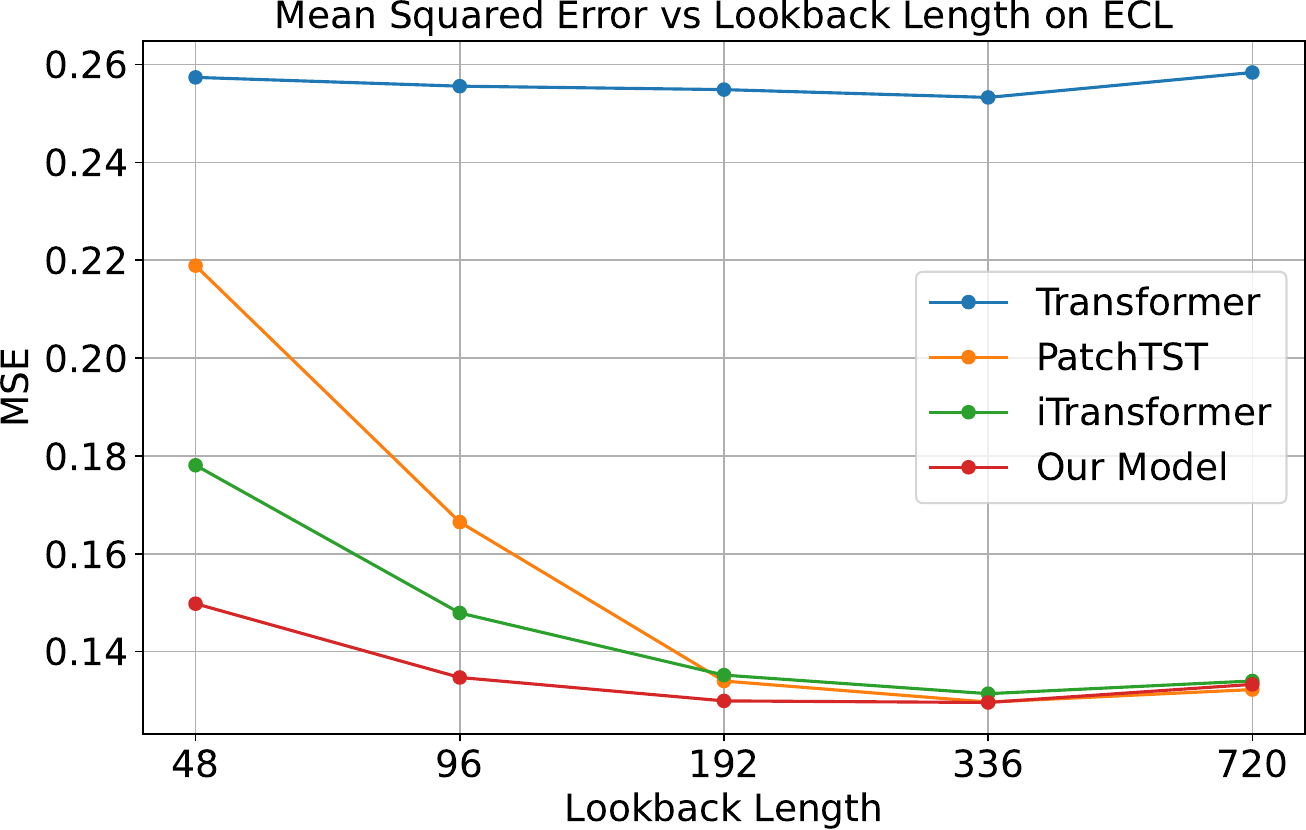}
        % \caption{TBD}
        % \label{fig:sub2}
    \end{subfigure}
    \caption{Performance with different lookback lengths and fixed prediction length $S = 96$. }
    \label{fig:lookback_lengh}
    \vspace{-2mm}
\end{figure}

% 1. Different lookback ({\color{blue} already got the results on Electricity and Weather Datasets}); For this, we want to show that, when we increase the lookback windows, the performance of our model will not decrease. That is usually the case for previous transformer-based model (before iTransformer). Plan to have three baselines: iTransformer, Transformer, PatchTST. Ideally, to show that our model outperforms others on all different lookback and the performance doesn't decrease with increasing lookback. 
\paragraph{The effect of different lookback lengths}
We also investigate how different lookback lengths would change the forecasting performance. 
With increased lookback lengths, we compare the forecasting performance of our model with that of several representative baselines in Figure~\ref{fig:lookback_lengh}. 
The results show that, when using a relatively short lookback length (i.e., 48), our model generally outperforms other models by a large margin.
It suggests that our model has a more powerful learning ability to capture the dependencies even with a short lookback length, while other models usually require longer lookback lengths to provide good performance. 
Moreover, by increasing the lookback length, the performances of our model and PatchTST usually improve, whereas the performance of Transformer remains almost the same on ECL dataset. 

\begin{wrapfigure}{r}{0.45\textwidth}
   \centering
    \includegraphics[width=\linewidth]{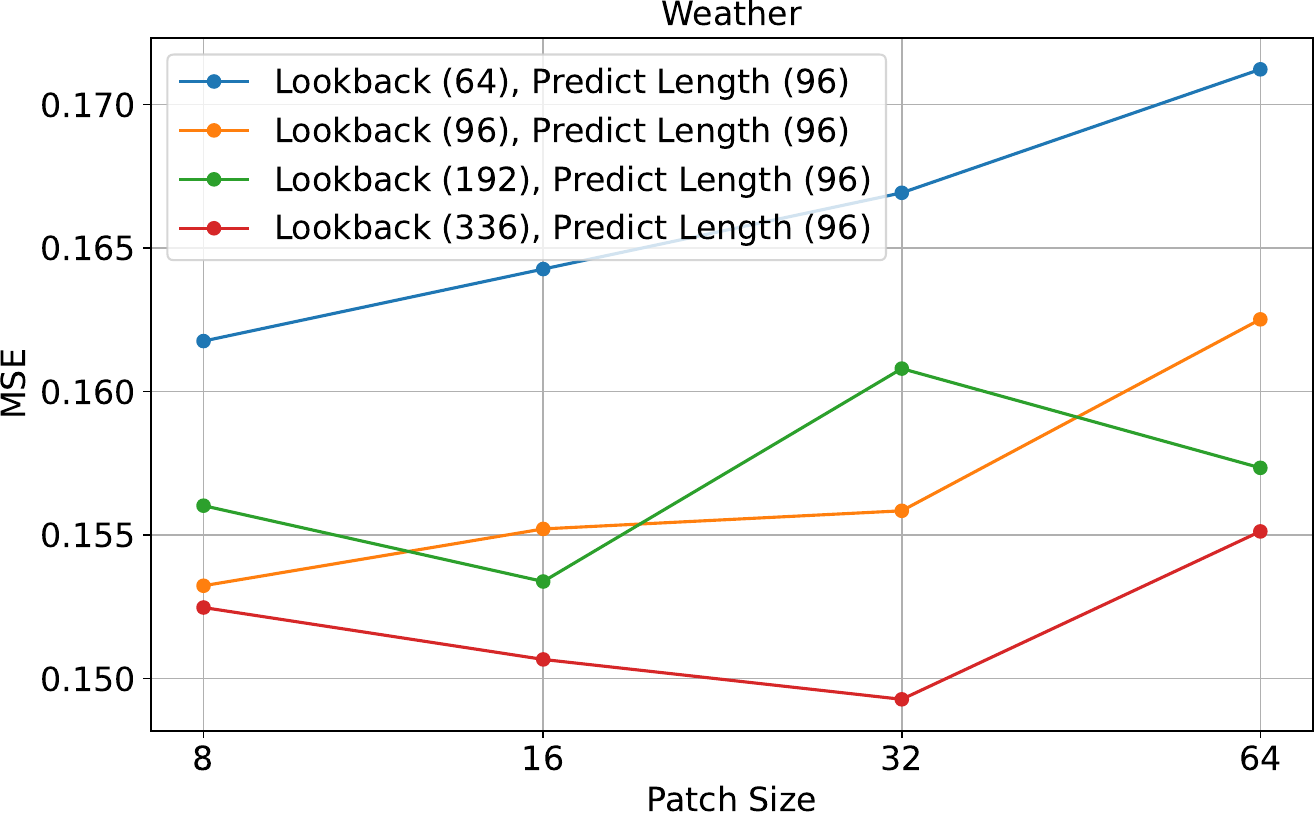}
    \caption{Performance with different patch sizes and lookback length.}
    \label{fig:patch_size}
\end{wrapfigure}

\paragraph{The effect of different patch sizes}
As we use patching in our model, we further examine the effect of different patch sizes. 
The patch size and the lookback length together determine the number of tokens for a variate. In Figure~\ref{fig:patch_size}, we demonstrate the performance by varying different patch sizes and lookback lengths. 
With lookback length of 64, the performance of using patch size 64 is much worse than that of patch size 8 It indicates that, when the number of tokens of a variate is extremely small (i.e., only 1 token for lookback length 64), the performance is not satisfactory as no enough fine-grained information. 
This could also be the reason why iTransformer may be not ideal in some cases - it use exactly a single token for a variate. 
Additionally, we also observe that, generally, for different lookback lengths, too small or too large patch size can lead to bad performance. 
The reason may be that too many tokens or too less tokens would increase the difficulty of training.

\paragraph{The number of dispatchers} 
In our model, we propose to use several dispatchers to reduce the memory complexity with the number of dispatchers as a hyper-parameter. 
Here, we dive deep into the tradeoff between GPU memory and MSE by varying the number of dispatchers. 
In Table~\ref{tab: vary_dispatchers}, we demonstrate the performance and GPU memory of different numbers of dispatchers on Weather and ECL with the prediction length as 96. 
The results show that, with only 5 dispatchers, the performance is usually worse than with more dispatchers. It suggests that we should avoid using too few dispatchers as it may affect the model performance. 
However, with fewer dispatchers, the GPU memory usage is less as shown in our complexity analysis in Section~\ref{sec: model_structure}. 
For larger datasets like ECL, increasing the number of dispatchers leads to more significant memory increase, compared with the smaller dataset (i.e., Weather). 

% 3. Vary the number of routers. This experiment is to show that how changing the number of routers would effect the performance. I would expect that within a reasonable range, the performance should be similar. But either too low or too high would make the performance worse. Too less routers may lead to "information squashing", too many routers may lead to no sufficient data to learn meaningful representations for the routers. 

% \usepackage{booktabs}
% \usepackage{multirow}
\begin{table}[h]
\centering
\caption{The performance and GPU memory usage of varying dispatchers on Weather and ECL.}
\label{tab: vary_dispatchers}
\begin{tabular}{@{}cc|cccc@{}}
\toprule
\multicolumn{2}{c|}{The number of dispatchers} & 5      & 10     & 20     & 50     \\ \midrule
\multirow{2}{*}{Weather}   & MSE               & 0.1575 & 0.1552 & 0.1573 & 0.1566 \\
                           & GPU Memory (GB)   & 2.165  & 2.191  & 2.233  & 2.405  \\ \midrule
\multirow{2}{*}{ECL}       & MSE               & 0.1348 & 0.1347 & 0.1343 & 0.1338 \\
                           & GPU Memory (GB)   & 12.807 & 13.389 & 14.335 & 16.509 \\ \bottomrule
\end{tabular}
\end{table}

\begin{figure}
    \centering
    \includegraphics[width=\linewidth]{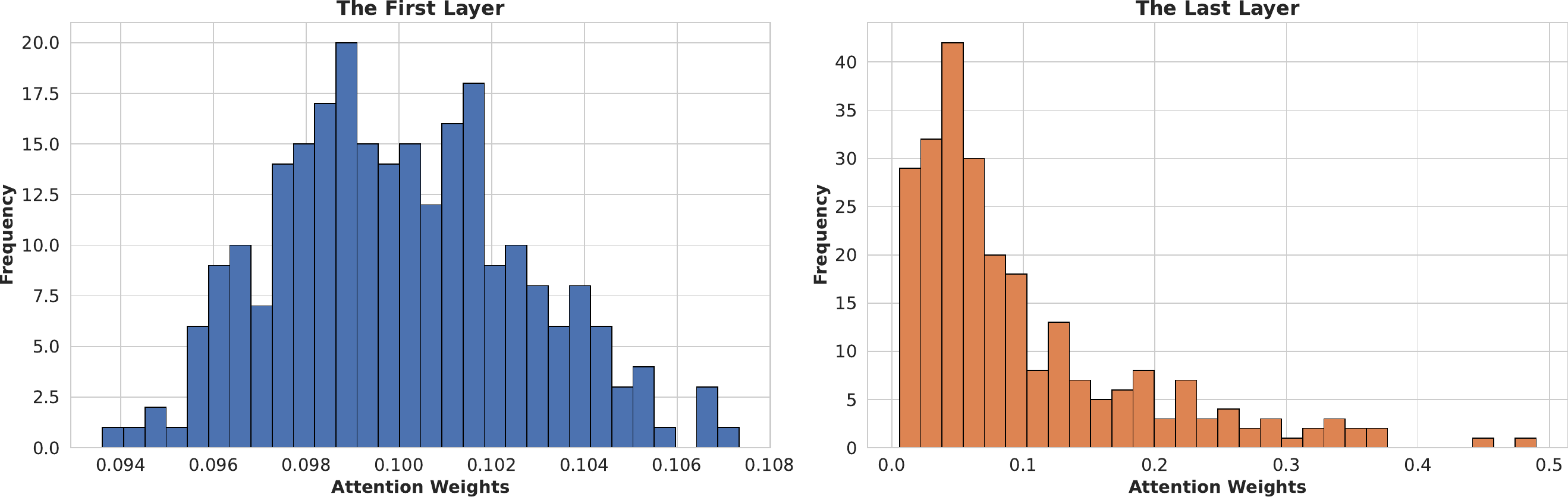}
    \caption{The distributions of multiplied attention weights between two patch tokens on Weather.}
    \label{fig: mlt_attn}
    \vspace{-2mm}
\end{figure}

\paragraph{Attention Weights}
With our dispatcher module, we have two attention weights matrices, one from patch tokens to dispatchers and one from dispatchers to patch tokens, with the size $N \times k$ and $k \times N$, respectively. Multiplying these two attention matrices gives us a new multiplied attention matrix with the size $N \times N$ that directly indicates the importance between two patch tokens. 
We demonstrate the multiplied attention weights from the first layer and the last layer in Figure~\ref{fig: mlt_attn}.
As shown, in the last layer, the distribution is visibly shifted to the left side, meaning that most of the token pairs have low attention weights, while a few token pairs have high attention weights. It may suggest that the last layer indeed learns how to distribute the information to important tokens. 
In contrast, the first layer has a more even distribution of attention weights, indicating that it distributes information more evenly to all tokens.

% \paragraph{Visualizations}, 
% Plan:
% 1. Attention Map vs data correlation in patch level (cross-channel).

% 2. (DONE) Attention weight histogram, to show that the histograms of different layers are quite different. And it learns something.
% \begin{figure}[ht]
%     \centering
%     \begin{subfigure}{\textwidth}
%         \centering
%         \includegraphics[width=\linewidth]{Figs/attn_token2dispatchers.pdf}
%         \caption{}
%         \label{fig:}
%     \end{subfigure}
%     % \hfill 
%     % \\
%     % \begin{subfigure}{\textwidth}
%     %     \centering
%     %     \includegraphics[width=\linewidth]{Figs/attn_dispatchers2token.pdf}
%     %     \caption{}
%     %     \label{fig:}
%     % \end{subfigure}
%     \\
%     \begin{subfigure}{\textwidth}
%     \centering
%     \includegraphics[width=\linewidth]{Figs/mlt_attn_token2token.pdf}
%     \caption{}
%     \label{fig:}
%     \end{subfigure}
%     \caption{\TODO{Just keep one, put the rest to the appendix}}
%     \label{fig: }
% \end{figure}

% 3. Given a patch, find 2 patches with high attention scores and 2 with low attention scores, Plot them and link them with the different attention weights. We would like to see that pairs with high attention scores share a more similar trend. 

% 4. Consider variate correlations, and compare the plots of variate correlations of ground truth with that of our predictions. To show that they are similar. Figure~\ref{fig:Variate Correlation Map} (\TODO{to make it smaller}). \TODO{Besides, that it would be better if we can provide that of iTransformer as well and show that it's worse. }
\begin{wrapfigure}{r}{0.45\textwidth}
   \centering
	\includegraphics[width=\linewidth]{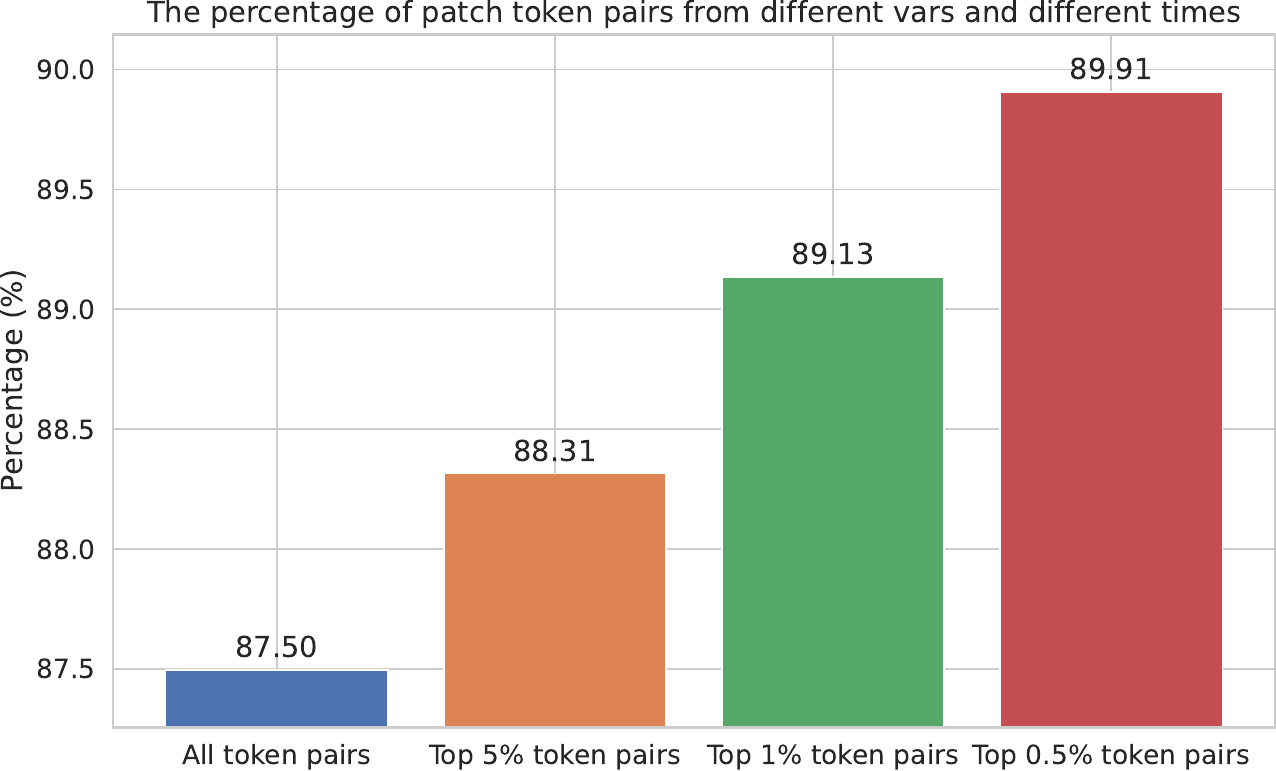}
	\caption{Patch token pairs with higher top attention weights are more likely from different variates and different times.}
	\label{fig: xvar_xtime_ratio}
    \vspace{-2.0mm}
\end{wrapfigure}
\paragraph{The importance of cross-variate cross-time dependencies}
With the multiplied attention weights, we further demonstrate the percentages of patch token pairs from different variables and different times for groups of patch tokens pairs with varied attention weights in Figure~\ref{fig: xvar_xtime_ratio}.
We observe that the groups of patch token pairs with higher attention weights have a higher percentage of pairs from different variates and different times. For example, for all token pairs, the percentage is 87.50, while the percentage is 89.91 for top 0.5\% token pairs with the highest attention weights. 
% It suggests that more important patch pairs are indeed more usually cross-variate cross-time. 
It suggests that more pairs of patch tokens with high attention weights come from different variates and times. 
Therefore, effectively modeling cross-variate cross-time is crucial for multivariate time series forecasting.

% \TODO{(DONE) Find some pairs of patches with top attention scores, and check how many pairs are from the same channel and how many of them are from different channels. If we can observe a significant ratio of pairs from different channels, we can highlight the importance of modeling cross-channel temporal dependencies.}

% \begin{figure}[ht]
%     \centering
%     \begin{subfigure}{0.4\textwidth}
%         \centering
%         \includegraphics[width=\linewidth]{Figs/Weather_ratio.pdf}
%         \caption{Weather}
%         \label{fig:sub1}
%     \end{subfigure}
%     % \hfill
%     \begin{subfigure}{0.4\textwidth}
%         \centering
%         \includegraphics[width=\linewidth]{Figs/Solar_ratio.pdf}
%         \caption{Solar}
%         \label{fig:sub2}
%     \end{subfigure}
%     \caption{The percentage of patch token pairs from the same variate.}
%     \label{fig: patch_pair_percentage}
% \end{figure}
\section{Conclusion and Future Work}
\label{sec:conclusion}
In this work, we first point out the limitation of previous works on time series transformers for multivariate forecasting: their lack of ability to effectively capture inter-series and intra-series dependencies simultaneously. 
We further demonstrate that inter-series and intra-series dependencies are crucial for multivariate time series forecasting as they commonly exist in real-world data. 
To mitigate this limitation of previous works, we propose a simple yet effective transformer model \ourmodel with a dispatcher mechanism to effectively capture inter-series and intra-series dependencies.
The experiments on 13 datasets for time series forecasting show that our model achieves superior performance compared with many representative baselines. 
Moreover, we conduct the ablation study and model analyses to verify the effectiveness of our dispatcher mechanism and demonstrate the importance of inter-series intra-series dependencies. 
Our study emphasizes the necessity and effectiveness of simultaneously capturing inter-variate and intra-variate dependencies in multivariate time series forecasting, and our proposed designs represent a step toward this goal. 

Although our model has the advantage of capturing inter-series and intra-series dependencies for multivariate time series data, our model may have a limitation in capturing these dependencies on extremely long time series due to the inherent limitation of Transformer architecture. 
How to enable time series Transformer to capture these dependencies with long lookback lengths and prediction lengths would be an interesting topic for future work. 

\bibliographystyle{plainnat}
\bibliography{neurips_2024}

\clearpage
\appendix
\appendixpage
\section{Diverse Cross-Time and Cross-Variate Dependencies}
\label{app:examples_diverse_dependencies}
\begin{figure}[h]
    \centering
    \begin{subfigure}{0.3\textwidth}
        \centering
        \includegraphics[width=\linewidth]{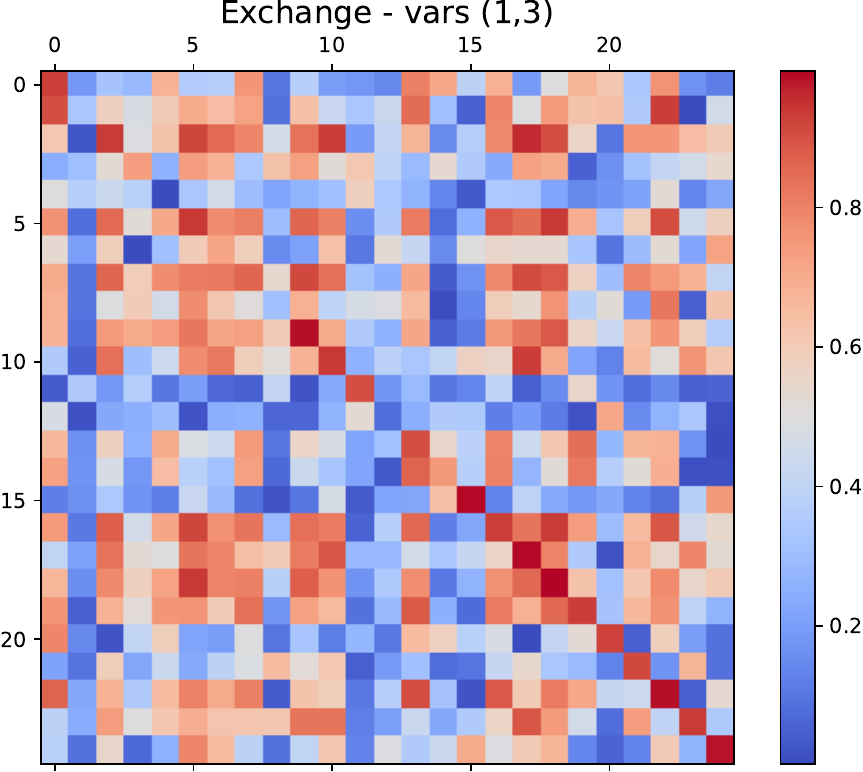}
        \caption{}
    \end{subfigure}
    \begin{subfigure}{0.3\textwidth}
        \centering
        \includegraphics[width=\linewidth]{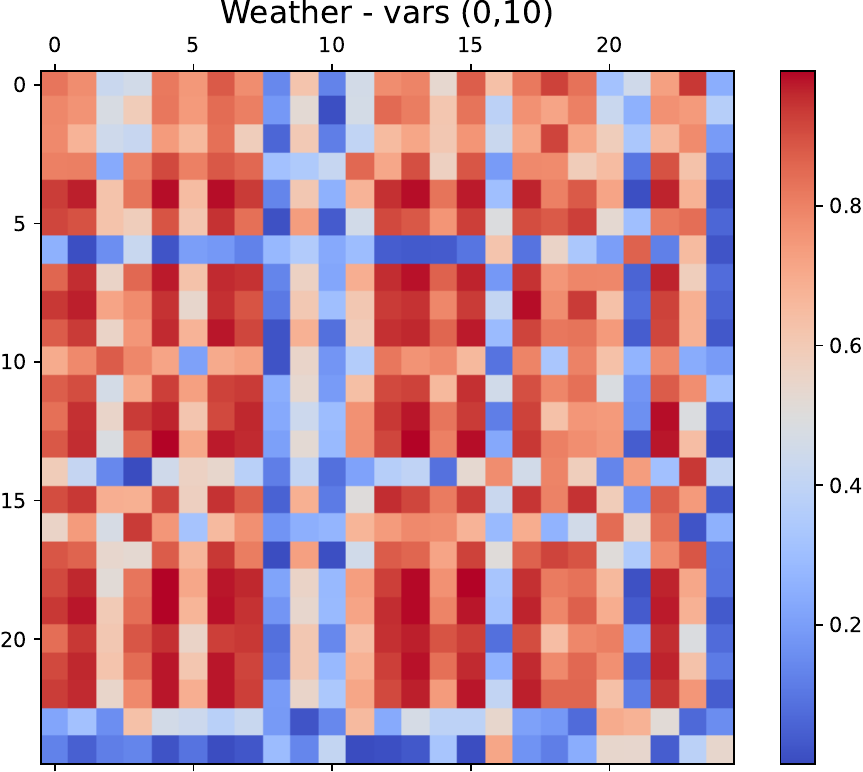}
        \caption{}
    \end{subfigure}
    \begin{subfigure}{0.3\textwidth}
        \centering
        \includegraphics[width=\linewidth]{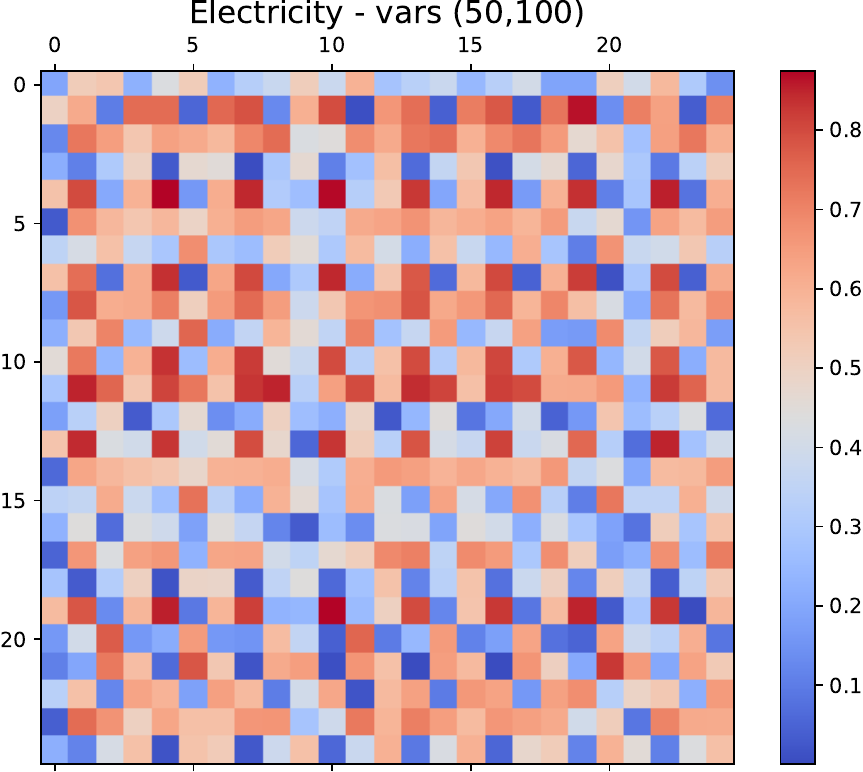}
        \caption{}
    \end{subfigure}
    \\
    \vfill
    \begin{subfigure}{0.3\textwidth}
    \centering
    \includegraphics[width=\linewidth]{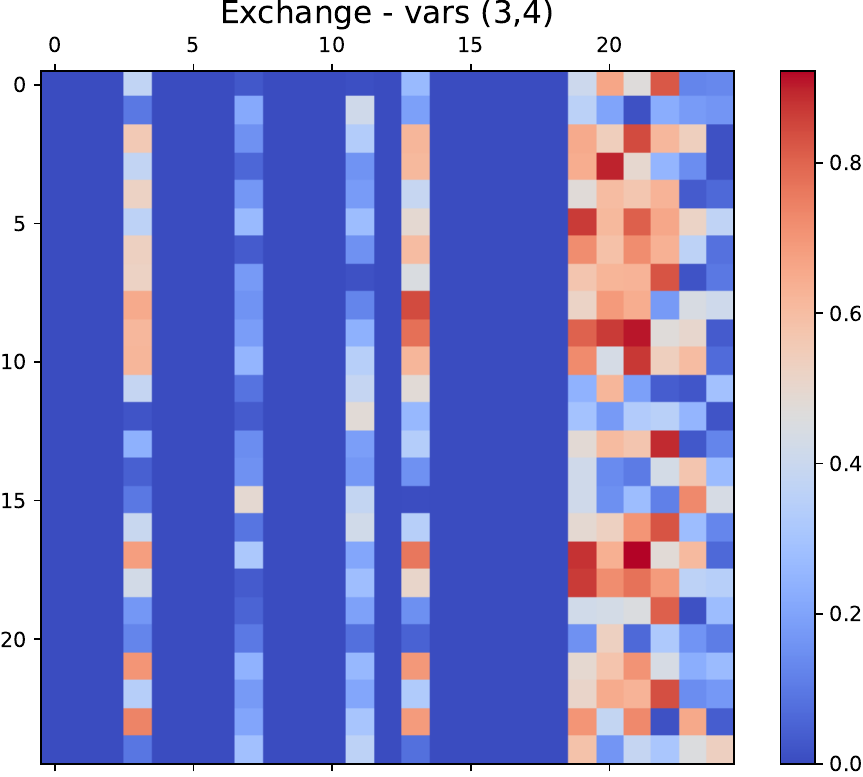}
    \caption{}
    \end{subfigure}
    \begin{subfigure}{0.3\textwidth}
        \centering
        \includegraphics[width=\linewidth]{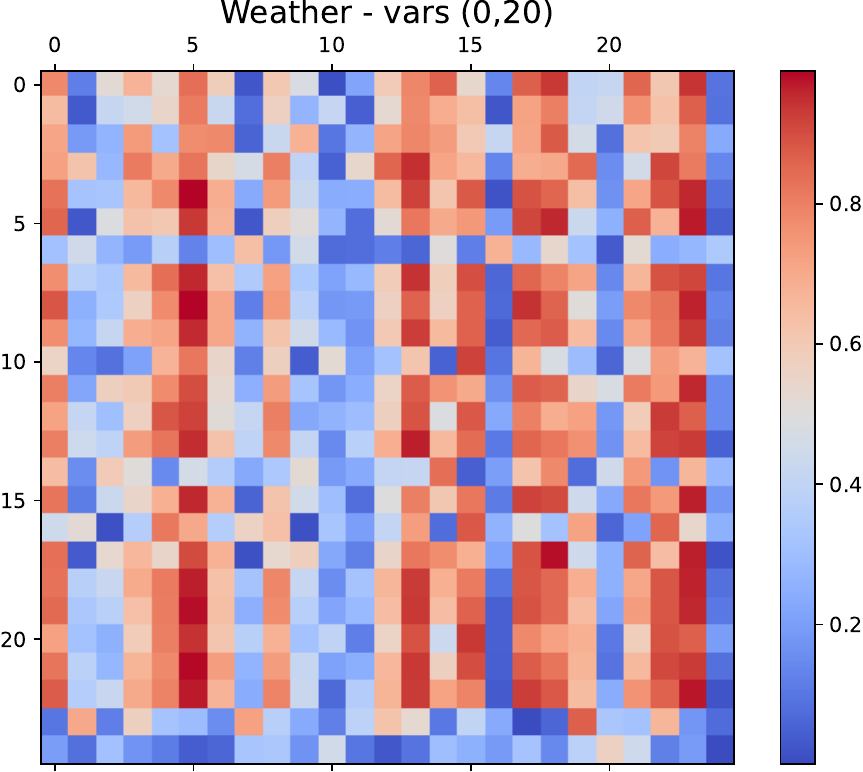}
        \caption{}
    \end{subfigure}
    \begin{subfigure}{0.3\textwidth}
        \centering
        \includegraphics[width=\linewidth]{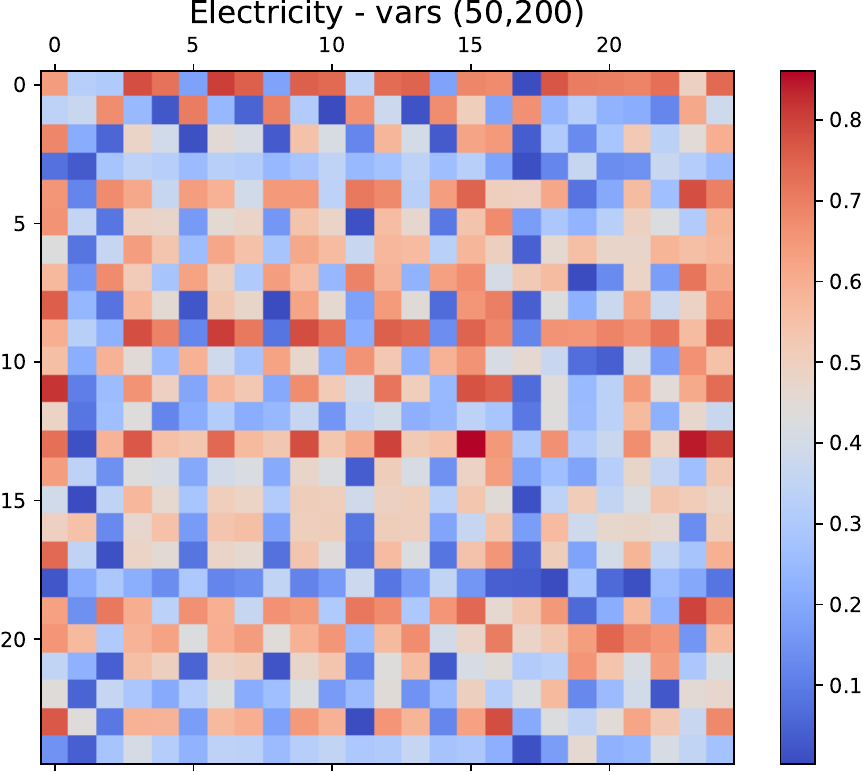}
        \caption{}
    \end{subfigure}
    \caption{Diverse cross-time cross-variate dependencies commonly exist in real-world data. }
    \label{fig:app_diverse_corr}
\end{figure}

We further illustrate the cross-time cross-variate correlations on Exchange, Weather, ECL datasets in Figure~\ref{fig:app_diverse_corr}.
We can see that correlation patterns for different datasets are quite different. Additionally, even for a specific dataset with different variate pairs, the correlations of cross-variate patch pairs are also very diverse.
For example, for Exchange, with variate pairs (1,3), the patches at the same time step are usually strongly correlated. In contrast, with variate pairs (3,4), the patches can sometimes even have zero correlation coefficient. 
Moreover, in Figure~\ref{fig:app_diverse_corr}, for a specific dataset with a specific pair of variates (i.e., in a subfigure), we have  similar observations as we discussed in Sec~\ref{sec: motivation} that there is no consistent correlation pattern for different patch pairs of two variates and inter-variate dependencies are at the fine-grained patch level. 
These examples further demonstrate the ubiquity and the diversity of these cross-time cross-variate correlations in real data. 
This also justifies the motivation of this paper -- propose a better method to explicitly model cross-time and cross-variate (intra-variate and inter-variate) dependencies. 
\section{More on Experiments}
\subsection{Datasets}
\label{app:datasets}
Following \citet{iTransformer}, 
we conduct experiments on 13 real-world datasets to evaluate the performance of our model including (1) a group of datasets -- ETT~\citep{Informer} contains 7 factors of electricity transformer from July 2016 to July 2018. There are four datasets where ETTm1 and ETTm2 are recorded every 15 minutes, and ETTh1 and ETTh2 are recorded every hour; (2) \update{Exchange~\citep{Autoformer} contains daily exchange rates from 8 countries from 1990 to 2016}. (3) Weather~\citep{Autoformer} collects the every 10-min data of  21 meteorological factors from the Weather Station of the Max Planck Biogeochemistry Institute in 2020. (4) ECL~\citep{Autoformer} records the electricity consumption data from 321 clients every hour. (5) Traffic~\citep{Autoformer} collects hourly road occupancy rates measured by 862 sensors of San Francisco Bay area freeways from January 2015 to December 2016. (6) Solar-Energy~\citep{LSTNet} records the solar power production of 137
PV plants in 2006, which are sampled every 10 minutes. (7) a group of datasets -- PEMS records the public traffic network data in California and collected by 5-minute windows. We use the same four public datasets (PEMS03, PEMS04, PEMS07, PEMS08) adopted in SCINet~\citep{SCINet} and iTransformer~\citep{iTransformer}.
We provide the detailed dataset statistics and descriptions in Table~\ref{tab:dataset}.

We also use the same train-validation-test splits as in TimesNet~\citep{Timesnet} and iTransformer~\citep{iTransformer}. For the forecasting setting, following iTansformer~\citep{iTransformer}, we use the fixed lookback length as 96 in all datasets.
In terms of the prediction lengths, we use the varied prediction lengths in \{96, 192, 336, 720\} for ETT, Exchange, Weather, ECL, Traffic, Solar-Energy. For PEMS datasets, we use the prediction lengths as \{12, 24, 48, 96\} for short-term forecasting.

\begin{table}[thbp]
  \vspace{0pt}
  \caption{\update{Detailed dataset statistics}. \emph{\# variates} denotes the variate number of each dataset. \emph{Dataset Size} denotes the total number of time points in (Train, Validation, Test) split respectively. \emph{Frequency} indicates the sampling interval of data points.}\label{tab:dataset}
  \vskip 0.05in
  \centering
  \begin{threeparttable}
  \begin{small}
  \renewcommand{\multirowsetup}{\centering}
  \setlength{\tabcolsep}{6.5pt}
  \begin{tabular}{l|c|c|c|c|c}
    \toprule
    Dataset Name & \# variates & Prediction Length & Dataset Size & Frequency& Information \\
    \toprule
     \update{ETTh1, ETTh2} & 7 & \scalebox{0.8}{\{96, 192, 336, 720\}} & (8545, 2881, 2881) & Hourly & Electricity\\
     \midrule
     \update{ETTm1, ETTm2} & 7 & \scalebox{0.8}{\{96, 192, 336, 720\}} & (34465, 11521, 11521) & 15min & Electricity\\
    \midrule
    \update{Exchange} & 8 & \scalebox{0.8}{\{96, 192, 336, 720\}} & (5120, 665, 1422) & Daily & Economy \\
    \midrule
    Weather & 21 & \scalebox{0.8}{\{96, 192, 336, 720\}} & (36792, 5271, 10540) & 10min & Weather\\
    \midrule
    ECL & 321 & \scalebox{0.8}{\{96, 192, 336, 720\}} & (18317, 2633, 5261) & Hourly & Electricity \\
    \midrule
    Traffic & 862 & \scalebox{0.8}{\{96, 192, 336, 720\}} & (12185, 1757, 3509) & Hourly & Transportation \\
    \midrule
    Solar-Energy & 137  & \scalebox{0.8}{\{96, 192, 336, 720\}} & (36601, 5161, 10417) & 10min & Energy \\
    \midrule
    \update{PEMS03} & 358 & \scalebox{0.8}{\{12, 24, 48, 96\}} & (15617, 5135, 5135) & 5min & Transportation\\
    \midrule
    \update{PEMS04} & 307 & \scalebox{0.8}{\{12, 24, 48, 96\}} & (10172, 3375, 3375) & 5min & Transportation\\
    \midrule
    \update{PEMS07} & 883 & \scalebox{0.8}{\{12, 24, 48, 96\}} & (16911, 5622, 5622) & 5min & Transportation\\
    \midrule
    \update{PEMS08} & 170 & \scalebox{0.8}{\{12, 24, 48, 96\}} & (10690, 3548, 3548) & 5min & Transportation \\
    \bottomrule
    \end{tabular}
    \end{small}
  \end{threeparttable}
  \vspace{-5pt}
\end{table}

\subsection{Experimental Setting}
\label{app: experimental_setting}
We conduct all the experiments with PyTorch~\citep{Pytorch} and utilize a single NVIDIA A100 GPU with 40GB memory.
We describe the hyperparameter choices used in our experiments in the following. 
For the optimizer, we use ADAM~\citep{Adam} with the learning rate in \{$10^{-3}$, $5\times 10^{-4}$, $10^{-4}$\}. 
The batch sizes are selected from \{16, 32, 64, 128\} depending on the dataset sizes. 
The maximum number of training epochs is set to 100 as in ~\citet{PatchTST}. Meanwhile, we also use the early stop strategy to stop the training when the loss does not decrease in 10 epochs. 
The number of layers of our Transformer blocks is selected from \{2,3,4\}. 
The hidden dimension of $D$ is set from \{128, 256, 512\}.

For the experimental results of our model, we report the averaged results with 5 runs with different seeds. 
For the results of previous models, we reuse the results from iTransformer paper~\citep{iTransformer} as we are using the same experimental setting. 

\subsection{Full Results of Forecasting}
Due to the space limitation, we only display the averaged results over 4 prediction lengths for datasets on long-term forecasting. Here, we provide the full results of long-term forecasting in Table~\ref{tab:full_baseline_results}. In summary, our model achieves the best results on 24 and 26 out of 36 settings with different prediction lengths among other baselines. 
\label{app:full_reults}
\begin{table}[htbp]
  \caption{\update{Full results of the long-term forecasting task}. We compare extensive competitive models under different prediction lengths following the setting of TimesNet~\citeyearpar{Timesnet}. The input sequence length is set to 96 for all baselines. \emph{Avg} means the average results from all four prediction lengths.}
  \label{tab:full_baseline_results}
  \vskip -0.0in
  \vspace{3pt}
  \renewcommand{\arraystretch}{0.85} 
  \centering
  \resizebox{1\columnwidth}{!}{
  \begin{threeparttable}
  \begin{small}
  \renewcommand{\multirowsetup}{\centering}
  \setlength{\tabcolsep}{1pt}
  \begin{tabular}{c|c|cc|cc|cc|cc|cc|cc|cc|cc|cc|cc|cc|cc}
    \toprule
    \multicolumn{2}{c}{\multirow{2}{*}{Models}} & 
    \multicolumn{2}{c}{\rotatebox{0}{\scalebox{0.8}{\textbf{\ourmodel}}}} &
    \multicolumn{2}{c}{\rotatebox{0}{\scalebox{0.8}{{iTransformer}}}} &
    \multicolumn{2}{c}{\rotatebox{0}{\scalebox{0.8}{\update{RLinear}}}} &
    \multicolumn{2}{c}{\rotatebox{0}{\scalebox{0.8}{PatchTST}}} &
    \multicolumn{2}{c}{\rotatebox{0}{\scalebox{0.8}{Crossformer}}}  &
    \multicolumn{2}{c}{\rotatebox{0}{\scalebox{0.8}{TiDE}}} &
    \multicolumn{2}{c}{\rotatebox{0}{\scalebox{0.8}{{TimesNet}}}} &
    \multicolumn{2}{c}{\rotatebox{0}{\scalebox{0.8}{DLinear}}}&
    \multicolumn{2}{c}{\rotatebox{0}{\scalebox{0.8}{SCINet}}} &
    \multicolumn{2}{c}{\rotatebox{0}{\scalebox{0.8}{FEDformer}}} &
    \multicolumn{2}{c}{\rotatebox{0}{\scalebox{0.8}{Stationary}}} &
    \multicolumn{2}{c}{\rotatebox{0}{\scalebox{0.8}{Autoformer}}} \\
    %&\multicolumn{2}{c}{\rotatebox{0}{\scalebox{0.8}{Informer}}} 
    \multicolumn{2}{c}{} &
    \multicolumn{2}{c}{\scalebox{0.8}{\textbf{(Ours)}}} & 
    \multicolumn{2}{c}{\scalebox{0.8}{\citeyearpar{li2023revisiting}}} & 
    \multicolumn{2}{c}{\scalebox{0.8}{\citeyearpar{li2023revisiting}}} & 
    \multicolumn{2}{c}{\scalebox{0.8}{\citeyearpar{PatchTST}}} & 
    \multicolumn{2}{c}{\scalebox{0.8}{\citeyearpar{Crossformer}}}  & 
    \multicolumn{2}{c}{\scalebox{0.8}{\citeyearpar{das2023long}}} & 
    \multicolumn{2}{c}{\scalebox{0.8}{\citeyearpar{Timesnet}}} & 
    \multicolumn{2}{c}{\scalebox{0.8}{\citeyearpar{DLinear}}}& 
    \multicolumn{2}{c}{\scalebox{0.8}{\citeyearpar{SCINet}}} &
    \multicolumn{2}{c}{\scalebox{0.8}{\citeyearpar{fedformer}}} &
    \multicolumn{2}{c}{\scalebox{0.8}{\citeyearpar{Stationary}}} &
    \multicolumn{2}{c}{\scalebox{0.8}{\citeyearpar{Autoformer}}} \\
    %&\multicolumn{2}{c}{\scalebox{0.8}{\citeyearpar{Informer}}} 
    \cmidrule(lr){3-4} \cmidrule(lr){5-6}\cmidrule(lr){7-8} \cmidrule(lr){9-10}\cmidrule(lr){11-12}\cmidrule(lr){13-14} \cmidrule(lr){15-16} \cmidrule(lr){17-18} \cmidrule(lr){19-20} \cmidrule(lr){21-22} \cmidrule(lr){23-24} \cmidrule(lr){25-26} 
    \multicolumn{2}{c}{Metric}  & \scalebox{0.78}{MSE} & \scalebox{0.78}{MAE}  & \scalebox{0.78}{MSE} & \scalebox{0.78}{MAE}  & \scalebox{0.78}{MSE} & \scalebox{0.78}{MAE}  & \scalebox{0.78}{MSE} & \scalebox{0.78}{MAE}  & \scalebox{0.78}{MSE} & \scalebox{0.78}{MAE}  & \scalebox{0.78}{MSE} & \scalebox{0.78}{MAE} & \scalebox{0.78}{MSE} & \scalebox{0.78}{MAE} & \scalebox{0.78}{MSE} & \scalebox{0.78}{MAE} & \scalebox{0.78}{MSE} & \scalebox{0.78}{MAE} & \scalebox{0.78}{MSE} & \scalebox{0.78}{MAE} & \scalebox{0.78}{MSE} & \scalebox{0.78}{MAE} & \scalebox{0.78}{MSE} & \scalebox{0.78}{MAE}\\
    \toprule
    
    \multirow{5}{*}{\update{\rotatebox{90}{\scalebox{0.95}{ETTm1}}}}
    &  \scalebox{0.78}{96} & \boldres{\scalebox{0.78}{0.313} }& \boldres{\scalebox{0.78}{0.352} }& {\scalebox{0.78}{0.334}} & {\scalebox{0.78}{0.368}} & \scalebox{0.78}{0.355} & \scalebox{0.78}{0.376} & \secondres{\scalebox{0.78}{0.329}} & \secondres{\scalebox{0.78}{0.367}} & \scalebox{0.78}{0.404} & \scalebox{0.78}{0.426} & \scalebox{0.78}{0.364} & \scalebox{0.78}{0.387} &{\scalebox{0.78}{0.338}} &{\scalebox{0.78}{0.375}} &{\scalebox{0.78}{0.345}} &{\scalebox{0.78}{0.372}} & \scalebox{0.78}{0.418} & \scalebox{0.78}{0.438} &\scalebox{0.78}{0.379} &\scalebox{0.78}{0.419} &\scalebox{0.78}{0.386} &\scalebox{0.78}{0.398} &\scalebox{0.78}{0.505} &\scalebox{0.78}{0.475} \\ %&\scalebox{0.78}{0.672} &\scalebox{0.78}{0.571} \\
    & \scalebox{0.78}{192} & \boldres{\scalebox{0.78}{0.359} }& \boldres{\scalebox{0.78}{0.380} }& \scalebox{0.78}{0.377} & \scalebox{0.78}{0.391} & \scalebox{0.78}{0.391} & \scalebox{0.78}{0.392} & \secondres{\scalebox{0.78}{0.367}} & \secondres{\scalebox{0.78}{0.385}} & \scalebox{0.78}{0.450} & \scalebox{0.78}{0.451} &\scalebox{0.78}{0.398} & \scalebox{0.78}{0.404} &{\scalebox{0.78}{0.374}} &{\scalebox{0.78}{0.387}}  &{\scalebox{0.78}{0.380}} &{\scalebox{0.78}{0.389}} & \scalebox{0.78}{0.439} & \scalebox{0.78}{0.450}  &\scalebox{0.78}{0.426} &\scalebox{0.78}{0.441} &\scalebox{0.78}{0.459} &\scalebox{0.78}{0.444} &\scalebox{0.78}{0.553} &\scalebox{0.78}{0.496} \\ %&\scalebox{0.78}{0.795} &\scalebox{0.78}{0.669}\\
    & \scalebox{0.78}{336} & \boldres{\scalebox{0.78}{0.395} }& \boldres{\scalebox{0.78}{0.404} }& \scalebox{0.78}{0.426} & \scalebox{0.78}{0.420} & \scalebox{0.78}{0.424} & \scalebox{0.78}{0.415} & \secondres{\scalebox{0.78}{0.399}} & \secondres{\scalebox{0.78}{0.410}} & \scalebox{0.78}{0.532}  &\scalebox{0.78}{0.515} & \scalebox{0.78}{0.428} & \scalebox{0.78}{0.425} &{\scalebox{0.78}{0.410}} &{\scalebox{0.78}{0.411}}  &{\scalebox{0.78}{0.413}} &{\scalebox{0.78}{0.413}} & \scalebox{0.78}{0.490} & \scalebox{0.78}{0.485}  &\scalebox{0.78}{0.445} &\scalebox{0.78}{0.459} &\scalebox{0.78}{0.495} &\scalebox{0.78}{0.464} &\scalebox{0.78}{0.621} &\scalebox{0.78}{0.537} \\ %&\scalebox{0.78}{1.212} &\scalebox{0.78}{0.871} \\
    & \scalebox{0.78}{720} & \boldres{\scalebox{0.78}{0.449} }& \secondres{\scalebox{0.78}{0.440} }& \scalebox{0.78}{0.491} & \scalebox{0.78}{0.459} & \scalebox{0.78}{0.487} & \scalebox{0.78}{0.450} & \secondres{\scalebox{0.78}{0.454}} & \boldres{\scalebox{0.78}{0.439}} & \scalebox{0.78}{0.666} & \scalebox{0.78}{0.589} & \scalebox{0.78}{0.487} & \scalebox{0.78}{0.461} &{\scalebox{0.78}{0.478}} &{\scalebox{0.78}{0.450}} &{\scalebox{0.78}{0.474}} &{\scalebox{0.78}{0.453}} & \scalebox{0.78}{0.595} & \scalebox{0.78}{0.550}  &\scalebox{0.78}{0.543} &\scalebox{0.78}{0.490} &\scalebox{0.78}{0.585} &\scalebox{0.78}{0.516} &\scalebox{0.78}{0.671} &\scalebox{0.78}{0.561} \\ %&\scalebox{0.78}{1.166} &\scalebox{0.78}{0.823} \\
    \cmidrule(lr){2-26}
    & \scalebox{0.78}{Avg} & \boldres{\scalebox{0.78}{0.379} }& \boldres{\scalebox{0.78}{0.394} }& \scalebox{0.78}{0.407} & \scalebox{0.78}{0.410} & \scalebox{0.78}{0.414} & \scalebox{0.78}{0.407} & \secondres{\scalebox{0.78}{0.387}} & \secondres{\scalebox{0.78}{0.400}} & \scalebox{0.78}{0.513} & \scalebox{0.78}{0.496} & \scalebox{0.78}{0.419} & \scalebox{0.78}{0.419} &{\scalebox{0.78}{0.400}} &{\scalebox{0.78}{0.406}}  &{\scalebox{0.78}{0.403}} &{\scalebox{0.78}{0.407}} & \scalebox{0.78}{0.485} & \scalebox{0.78}{0.481}  &\scalebox{0.78}{0.448} &\scalebox{0.78}{0.452} &\scalebox{0.78}{0.481} &\scalebox{0.78}{0.456} &\scalebox{0.78}{0.588} &\scalebox{0.78}{0.517} \\ %&\scalebox{0.78}{0.961} &\scalebox{0.78}{0.734} \\
    \midrule
    
    \multirow{5}{*}{\update{\rotatebox{90}{\scalebox{0.95}{ETTm2}}}}
    &  \scalebox{0.78}{96} & \secondres{\scalebox{0.78}{0.178} }& \secondres{\scalebox{0.78}{0.262} }& {\scalebox{0.78}{0.180}} & {\scalebox{0.78}{0.264}} & \scalebox{0.78}{0.182} & \scalebox{0.78}{0.265} & \boldres{\scalebox{0.78}{0.175}} & \boldres{\scalebox{0.78}{0.259}} & \scalebox{0.78}{0.287} & \scalebox{0.78}{0.366} & \scalebox{0.78}{0.207} & \scalebox{0.78}{0.305} &{\scalebox{0.78}{0.187}} &\scalebox{0.78}{0.267} &\scalebox{0.78}{0.193} &\scalebox{0.78}{0.292} & \scalebox{0.78}{0.286} & \scalebox{0.78}{0.377} &\scalebox{0.78}{0.203} &\scalebox{0.78}{0.287} &{\scalebox{0.78}{0.192}} &\scalebox{0.78}{0.274} &\scalebox{0.78}{0.255} &\scalebox{0.78}{0.339} \\ % &\scalebox{0.78}{0.365} &\scalebox{0.78}{0.453} \\
    & \scalebox{0.78}{192} & \secondres{\scalebox{0.78}{0.243} }& \secondres{\scalebox{0.78}{0.304} }& \scalebox{0.78}{0.250} & {\scalebox{0.78}{0.309}} & {\scalebox{0.78}{0.246}} & {\scalebox{0.78}{0.304}} & \boldres{\scalebox{0.78}{0.241}} & \boldres{\scalebox{0.78}{0.302}} & \scalebox{0.78}{0.414} & \scalebox{0.78}{0.492} & \scalebox{0.78}{0.290} & \scalebox{0.78}{0.364} &{\scalebox{0.78}{0.249}} &{\scalebox{0.78}{0.309}} &\scalebox{0.78}{0.284} &\scalebox{0.78}{0.362} & \scalebox{0.78}{0.399} & \scalebox{0.78}{0.445} &\scalebox{0.78}{0.269} &\scalebox{0.78}{0.328} &\scalebox{0.78}{0.280} &\scalebox{0.78}{0.339} &\scalebox{0.78}{0.281} &\scalebox{0.78}{0.340} \\ %&\scalebox{0.78}{0.533} &\scalebox{0.78}{0.563} \\
    & \scalebox{0.78}{336} & \boldres{\scalebox{0.78}{0.302} }& \boldres{\scalebox{0.78}{0.341} }& {\scalebox{0.78}{0.311}} & {\scalebox{0.78}{0.348}} & {\scalebox{0.78}{0.307}} & \secondres{\scalebox{0.78}{0.342}} & \secondres{\scalebox{0.78}{0.305}} & {\scalebox{0.78}{0.343}}  & \scalebox{0.78}{0.597} & \scalebox{0.78}{0.542}  & \scalebox{0.78}{0.377} & \scalebox{0.78}{0.422} &{\scalebox{0.78}{0.321}} &{\scalebox{0.78}{0.351}} &\scalebox{0.78}{0.369} &\scalebox{0.78}{0.427} & \scalebox{0.78}{0.637} & \scalebox{0.78}{0.591} &\scalebox{0.78}{0.325} &\scalebox{0.78}{0.366} &\scalebox{0.78}{0.334} &\scalebox{0.78}{0.361} &\scalebox{0.78}{0.339} &\scalebox{0.78}{0.372} \\ %&\scalebox{0.78}{1.363} &\scalebox{0.78}{0.887} \\
    & \scalebox{0.78}{720} & \boldres{\scalebox{0.78}{0.398} }& \boldres{\scalebox{0.78}{0.395} }& \scalebox{0.78}{0.412} & \scalebox{0.78}{0.407} & {\scalebox{0.78}{0.407}} & \secondres{\scalebox{0.78}{0.398}} & \secondres{\scalebox{0.78}{0.402}} & {\scalebox{0.78}{0.400}} & \scalebox{0.78}{1.730} & \scalebox{0.78}{1.042} & \scalebox{0.78}{0.558} & \scalebox{0.78}{0.524} &{\scalebox{0.78}{0.408}} &{\scalebox{0.78}{0.403}} &\scalebox{0.78}{0.554} &\scalebox{0.78}{0.522} & \scalebox{0.78}{0.960} & \scalebox{0.78}{0.735} &\scalebox{0.78}{0.421} &\scalebox{0.78}{0.415} &\scalebox{0.78}{0.417} &\scalebox{0.78}{0.413} &\scalebox{0.78}{0.433} &\scalebox{0.78}{0.432} \\ %&\scalebox{0.78}{3.379} &\scalebox{0.78}{1.338} \\
    \cmidrule(lr){2-26}
    & \scalebox{0.78}{Avg} & \boldres{\scalebox{0.78}{0.280} }& \boldres{\scalebox{0.78}{0.326} }& {\scalebox{0.78}{0.288}} & {\scalebox{0.78}{0.332}} & {\scalebox{0.78}{0.286}} & {\scalebox{0.78}{0.327}} & \secondres{\scalebox{0.78}{0.281}} & \boldres{\scalebox{0.78}{0.326}} & \scalebox{0.78}{0.757} & \scalebox{0.78}{0.610} & \scalebox{0.78}{0.358} & \scalebox{0.78}{0.404} &{\scalebox{0.78}{0.291}} &{\scalebox{0.78}{0.333}} &\scalebox{0.78}{0.350} &\scalebox{0.78}{0.401} & \scalebox{0.78}{0.571} & \scalebox{0.78}{0.537} &\scalebox{0.78}{0.305} &\scalebox{0.78}{0.349} &\scalebox{0.78}{0.306} &\scalebox{0.78}{0.347} &\scalebox{0.78}{0.327} &\scalebox{0.78}{0.371} \\ %&\scalebox{0.78}{1.410} &\scalebox{0.78}{0.810} \\
    \midrule
    
    \multirow{5}{*}{\rotatebox{90}{\update{\scalebox{0.95}{ETTh1}}}}
    &  \scalebox{0.78}{96} & \secondres{\scalebox{0.78}{0.383} }& \secondres{\scalebox{0.78}{0.398} }& {\scalebox{0.78}{0.386}} & {\scalebox{0.78}{0.405}} & \scalebox{0.78}{0.386} & \boldres{\scalebox{0.78}{0.395}} & \scalebox{0.78}{0.414} & \scalebox{0.78}{0.419} & \scalebox{0.78}{0.423} & \scalebox{0.78}{0.448} & \scalebox{0.78}{0.479}& \scalebox{0.78}{0.464}  &{\scalebox{0.78}{0.384}} &{\scalebox{0.78}{0.402}} & \scalebox{0.78}{0.386} &{\scalebox{0.78}{0.400}} & \scalebox{0.78}{0.654} & \scalebox{0.78}{0.599} &\boldres{\scalebox{0.78}{0.376}} &\scalebox{0.78}{0.419} &\scalebox{0.78}{0.513} &\scalebox{0.78}{0.491} &\scalebox{0.78}{0.449} &\scalebox{0.78}{0.459}  \\ %&\scalebox{0.78}{0.865} &\scalebox{0.78}{0.713} \\
    & \scalebox{0.78}{192} & \secondres{\scalebox{0.78}{0.434} }& \secondres{\scalebox{0.78}{0.426} }& \scalebox{0.78}{0.441} & \scalebox{0.78}{0.436} & {\scalebox{0.78}{0.437}} & \boldres{\scalebox{0.78}{0.424}} & \scalebox{0.78}{0.460} & \scalebox{0.78}{0.445} & \scalebox{0.78}{0.471} & \scalebox{0.78}{0.474}  & \scalebox{0.78}{0.525} & \scalebox{0.78}{0.492} &{\scalebox{0.78}{0.436}} &{\scalebox{0.78}{0.429}}  &{\scalebox{0.78}{0.437}} &{\scalebox{0.78}{0.432}} & \scalebox{0.78}{0.719} & \scalebox{0.78}{0.631} &\boldres{\scalebox{0.78}{0.420}} &\scalebox{0.78}{0.448} &\scalebox{0.78}{0.534} &\scalebox{0.78}{0.504} &\scalebox{0.78}{0.500} &\scalebox{0.78}{0.482} \\ %&\scalebox{0.78}{1.008} &\scalebox{0.78}{0.792} \\
    & \scalebox{0.78}{336} & \secondres{\scalebox{0.78}{0.471} }& \boldres{\scalebox{0.78}{0.445} }& {\scalebox{0.78}{0.487}} & {\scalebox{0.78}{0.458}} & {\scalebox{0.78}{0.479}} & \secondres{\scalebox{0.78}{0.446}} & \scalebox{0.78}{0.501} & \scalebox{0.78}{0.466} & \scalebox{0.78}{0.570} & \scalebox{0.78}{0.546} & \scalebox{0.78}{0.565} & \scalebox{0.78}{0.515} &\scalebox{0.78}{0.491} &\scalebox{0.78}{0.469} &{\scalebox{0.78}{0.481}} & {\scalebox{0.78}{0.459}} & \scalebox{0.78}{0.778} & \scalebox{0.78}{0.659} &\boldres{\scalebox{0.78}{0.459}} &{\scalebox{0.78}{0.465}} &\scalebox{0.78}{0.588} &\scalebox{0.78}{0.535} &\scalebox{0.78}{0.521} &\scalebox{0.78}{0.496} \\ %&\scalebox{0.78}{1.107} &\scalebox{0.78}{0.809} \\
    & \scalebox{0.78}{720} & \boldres{\scalebox{0.78}{0.479} }& \boldres{\scalebox{0.78}{0.469} }& {\scalebox{0.78}{0.503}} & {\scalebox{0.78}{0.491}} & \secondres{\scalebox{0.78}{0.481}} & 
    \secondres{\scalebox{0.78}{0.470}} & {\scalebox{0.78}{0.500}} & {\scalebox{0.78}{0.488}} & \scalebox{0.78}{0.653} & \scalebox{0.78}{0.621} & \scalebox{0.78}{0.594} & \scalebox{0.78}{0.558} &\scalebox{0.78}{0.521} &{\scalebox{0.78}{0.500}} &\scalebox{0.78}{0.519} &\scalebox{0.78}{0.516} & \scalebox{0.78}{0.836} & \scalebox{0.78}{0.699} &{\scalebox{0.78}{0.506}} &{\scalebox{0.78}{0.507}} &\scalebox{0.78}{0.643} &\scalebox{0.78}{0.616} &{\scalebox{0.78}{0.514}} &\scalebox{0.78}{0.512}  \\ %&\scalebox{0.78}{1.181} &\scalebox{0.78}{0.865} \\
    \cmidrule(lr){2-26}
    & \scalebox{0.78}{Avg} & \secondres{\scalebox{0.78}{0.442} }& \secondres{\scalebox{0.78}{0.435} }& {\scalebox{0.78}{0.454}} & {\scalebox{0.78}{0.447}} & {\scalebox{0.78}{0.446}} & \boldres{\scalebox{0.78}{0.434}} & \scalebox{0.78}{0.469} & \scalebox{0.78}{0.454} & \scalebox{0.78}{0.529} & \scalebox{0.78}{0.522} & \scalebox{0.78}{0.541} & \scalebox{0.78}{0.507} &\scalebox{0.78}{0.458} &{\scalebox{0.78}{0.450}} &{\scalebox{0.78}{0.456}} &{\scalebox{0.78}{0.452}} & \scalebox{0.78}{0.747} & \scalebox{0.78}{0.647} &\boldres{\scalebox{0.78}{0.440}} &\scalebox{0.78}{0.460} &\scalebox{0.78}{0.570} &\scalebox{0.78}{0.537} &\scalebox{0.78}{0.496} &\scalebox{0.78}{0.487}  \\ %&\scalebox{0.78}{1.040} &\scalebox{0.78}{0.795} \\
    \midrule

    \multirow{5}{*}{\rotatebox{90}{\scalebox{0.95}{ETTh2}}}
    &  \scalebox{0.78}{96} & \secondres{\scalebox{0.78}{0.292} }& \secondres{\scalebox{0.78}{0.342} }& {\scalebox{0.78}{0.297}} & {\scalebox{0.78}{0.349}} & \boldres{\scalebox{0.78}{0.288}} & \boldres{\scalebox{0.78}{0.338}} & {\scalebox{0.78}{0.302}} & {\scalebox{0.78}{0.348}} & \scalebox{0.78}{0.745} & \scalebox{0.78}{0.584} &\scalebox{0.78}{0.400} & \scalebox{0.78}{0.440}  & {\scalebox{0.78}{0.340}} & {\scalebox{0.78}{0.374}} &{\scalebox{0.78}{0.333}} &{\scalebox{0.78}{0.387}} & \scalebox{0.78}{0.707} & \scalebox{0.78}{0.621}  &\scalebox{0.78}{0.358} &\scalebox{0.78}{0.397} &\scalebox{0.78}{0.476} &\scalebox{0.78}{0.458} &\scalebox{0.78}{0.346} &\scalebox{0.78}{0.388} \\ %&\scalebox{0.78}{3.755} &\scalebox{0.78}{1.525} \\
    & \scalebox{0.78}{192} & \boldres{\scalebox{0.78}{0.370} }& \boldres{\scalebox{0.78}{0.390} }& {\scalebox{0.78}{0.380}} & {\scalebox{0.78}{0.400}} & \secondres{\scalebox{0.78}{0.374}} & \boldres{\scalebox{0.78}{0.390}} &{\scalebox{0.78}{0.388}} & {\scalebox{0.78}{0.400}} & \scalebox{0.78}{0.877} & \scalebox{0.78}{0.656} & \scalebox{0.78}{0.528} & \scalebox{0.78}{0.509} & {\scalebox{0.78}{0.402}} & {\scalebox{0.78}{0.414}} &\scalebox{0.78}{0.477} &\scalebox{0.78}{0.476} & \scalebox{0.78}{0.860} & \scalebox{0.78}{0.689} &{\scalebox{0.78}{0.429}} &{\scalebox{0.78}{0.439}} &\scalebox{0.78}{0.512} &\scalebox{0.78}{0.493} &\scalebox{0.78}{0.456} &\scalebox{0.78}{0.452} \\ %&\scalebox{0.78}{5.602} &\scalebox{0.78}{1.931} \\
    & \scalebox{0.78}{336} & \boldres{\scalebox{0.78}{0.382} }& \boldres{\scalebox{0.78}{0.408} }& {\scalebox{0.78}{0.428}} & {\scalebox{0.78}{0.432}} & \secondres{\scalebox{0.78}{0.415}} & \secondres{\scalebox{0.78}{0.426}} & {\scalebox{0.78}{0.426}} & {\scalebox{0.78}{0.433}}& \scalebox{0.78}{1.043} & \scalebox{0.78}{0.731} & \scalebox{0.78}{0.643} & \scalebox{0.78}{0.571}  & {\scalebox{0.78}{0.452}} & {\scalebox{0.78}{0.452}} &\scalebox{0.78}{0.594} &\scalebox{0.78}{0.541} & \scalebox{0.78}{1.000} &\scalebox{0.78}{0.744} &\scalebox{0.78}{0.496} &\scalebox{0.78}{0.487} &\scalebox{0.78}{0.552} &\scalebox{0.78}{0.551} &{\scalebox{0.78}{0.482}} &\scalebox{0.78}{0.486}\\ % &\scalebox{0.78}{4.721} &\scalebox{0.78}{1.835} \\
    & \scalebox{0.78}{720} & \boldres{\scalebox{0.78}{0.409} }& \boldres{\scalebox{0.78}{0.431} }& {\scalebox{0.78}{0.427}} & {\scalebox{0.78}{0.445}} & \secondres{\scalebox{0.78}{0.420}} & \secondres{\scalebox{0.78}{0.440}} & {\scalebox{0.78}{0.431}} & {\scalebox{0.78}{0.446}} & \scalebox{0.78}{1.104} & \scalebox{0.78}{0.763} & \scalebox{0.78}{0.874} & \scalebox{0.78}{0.679} & {\scalebox{0.78}{0.462}} & {\scalebox{0.78}{0.468}} &\scalebox{0.78}{0.831} &\scalebox{0.78}{0.657} & \scalebox{0.78}{1.249} & \scalebox{0.78}{0.838} &{\scalebox{0.78}{0.463}} &{\scalebox{0.78}{0.474}} &\scalebox{0.78}{0.562} &\scalebox{0.78}{0.560} &\scalebox{0.78}{0.515} &\scalebox{0.78}{0.511} \\ %&\scalebox{0.78}{3.647} &\scalebox{0.78}{1.625} \\
    \cmidrule(lr){2-26}
    & \scalebox{0.78}{Avg} & \boldres{\scalebox{0.78}{0.363} }& \boldres{\scalebox{0.78}{0.393} }& {\scalebox{0.78}{0.383}} & {\scalebox{0.78}{0.407}} & \secondres{\scalebox{0.78}{0.374}} & \secondres{\scalebox{0.78}{0.398}} & {\scalebox{0.78}{0.387}} & {\scalebox{0.78}{0.407}} & \scalebox{0.78}{0.942} & \scalebox{0.78}{0.684} & \scalebox{0.78}{0.611} & \scalebox{0.78}{0.550}  &{\scalebox{0.78}{0.414}} &{\scalebox{0.78}{0.427}} &\scalebox{0.78}{0.559} &\scalebox{0.78}{0.515} & \scalebox{0.78}{0.954} & \scalebox{0.78}{0.723} &\scalebox{0.78}{{0.437}} &\scalebox{0.78}{{0.449}} &\scalebox{0.78}{0.526} &\scalebox{0.78}{0.516} &\scalebox{0.78}{0.450} &\scalebox{0.78}{0.459} \\ %&\scalebox{0.78}{4.431} &\scalebox{0.78}{1.729} \\
    \midrule
    
    \multirow{5}{*}{\rotatebox{90}{\scalebox{0.95}{ECL}}} 
    &  \scalebox{0.78}{96} & \boldres{\scalebox{0.78}{0.139} }& \boldres{\scalebox{0.78}{0.235} }& \secondres{\scalebox{0.78}{0.148}} & \secondres{\scalebox{0.78}{0.240}} & \scalebox{0.78}{0.201} & \scalebox{0.78}{0.281} & \scalebox{0.78}{0.181} & {\scalebox{0.78}{0.270}} & \scalebox{0.78}{0.219} & \scalebox{0.78}{0.314} & \scalebox{0.78}{0.237} & \scalebox{0.78}{0.329} &{\scalebox{0.78}{0.168}} &\scalebox{0.78}{0.272} &\scalebox{0.78}{0.197} &\scalebox{0.78}{0.282} & \scalebox{0.78}{0.247} & \scalebox{0.78}{0.345} &\scalebox{0.78}{0.193} &\scalebox{0.78}{0.308} &{\scalebox{0.78}{0.169}} &{\scalebox{0.78}{0.273}} &\scalebox{0.78}{0.201} &\scalebox{0.78}{0.317}  \\ %&\scalebox{0.78}{0.274} &\scalebox{0.78}{0.368} \\
    & \scalebox{0.78}{192} & \boldres{\scalebox{0.78}{0.155} }& \boldres{\scalebox{0.78}{0.250} }& \secondres{\scalebox{0.78}{0.162}} & \secondres{\scalebox{0.78}{0.253}} & \scalebox{0.78}{0.201} & \scalebox{0.78}{0.283} & \scalebox{0.78}{0.188} & {\scalebox{0.78}{0.274}} & \scalebox{0.78}{0.231} & \scalebox{0.78}{0.322} & \scalebox{0.78}{0.236} & \scalebox{0.78}{0.330} &{\scalebox{0.78}{0.184}} &\scalebox{0.78}{0.289} &\scalebox{0.78}{0.196} &{\scalebox{0.78}{0.285}} & \scalebox{0.78}{0.257} & \scalebox{0.78}{0.355} &\scalebox{0.78}{0.201} &\scalebox{0.78}{0.315} &{\scalebox{0.78}{0.182}} &\scalebox{0.78}{0.286} &\scalebox{0.78}{0.222} &\scalebox{0.78}{0.334} \\ %&\scalebox{0.78}{0.296} &\scalebox{0.78}{0.386} \\
    & \scalebox{0.78}{336} & \boldres{\scalebox{0.78}{0.170} }& \boldres{\scalebox{0.78}{0.268} }& \secondres{\scalebox{0.78}{0.178}} & \secondres{\scalebox{0.78}{0.269}} & \scalebox{0.78}{0.215} & \scalebox{0.78}{0.298} & \scalebox{0.78}{0.204} & {\scalebox{0.78}{0.293}} & \scalebox{0.78}{0.246} & \scalebox{0.78}{0.337} & \scalebox{0.78}{0.249} & \scalebox{0.78}{0.344} &{\scalebox{0.78}{0.198}} &{\scalebox{0.78}{0.300}} &\scalebox{0.78}{0.209} &{\scalebox{0.78}{0.301}} & \scalebox{0.78}{0.269} & \scalebox{0.78}{0.369} &\scalebox{0.78}{0.214} &\scalebox{0.78}{0.329} &{\scalebox{0.78}{0.200}} &\scalebox{0.78}{0.304} &\scalebox{0.78}{0.231} &\scalebox{0.78}{0.338}  \\ %&\scalebox{0.78}{0.300} &\scalebox{0.78}{0.394} \\
    & \scalebox{0.78}{720} & \boldres{\scalebox{0.78}{0.198} }& \boldres{\scalebox{0.78}{0.293} }& {\scalebox{0.78}{0.225}} & \secondres{\scalebox{0.78}{0.317}} & \scalebox{0.78}{0.257} & \scalebox{0.78}{0.331} & \scalebox{0.78}{0.246} & \scalebox{0.78}{0.324} & \scalebox{0.78}{0.280} & \scalebox{0.78}{0.363} & \scalebox{0.78}{0.284} & \scalebox{0.78}{0.373} &\secondres{\scalebox{0.78}{0.220}} &{\scalebox{0.78}{0.320}} &\scalebox{0.78}{0.245} &\scalebox{0.78}{0.333} & \scalebox{0.78}{0.299} & \scalebox{0.78}{0.390} &\scalebox{0.78}{0.246} &\scalebox{0.78}{0.355} &{\scalebox{0.78}{0.222}} &{\scalebox{0.78}{0.321}} &\scalebox{0.78}{0.254} &\scalebox{0.78}{0.361} \\ %&\scalebox{0.78}{0.373} &\scalebox{0.78}{0.439} \\
    \cmidrule(lr){2-26}
    & \scalebox{0.78}{Avg} & \boldres{\scalebox{0.78}{0.166} }& \boldres{\scalebox{0.78}{0.262} }& \secondres{\scalebox{0.78}{0.178}} & \secondres{\scalebox{0.78}{0.270}} & \scalebox{0.78}{0.219} & \scalebox{0.78}{0.298} & \scalebox{0.78}{0.205} & {\scalebox{0.78}{0.290}} & \scalebox{0.78}{0.244} & \scalebox{0.78}{0.334} & \scalebox{0.78}{0.251} & \scalebox{0.78}{0.344} &{\scalebox{0.78}{0.192}} &\scalebox{0.78}{0.295} &\scalebox{0.78}{0.212} &\scalebox{0.78}{0.300} & \scalebox{0.78}{0.268} & \scalebox{0.78}{0.365} &\scalebox{0.78}{0.214} &\scalebox{0.78}{0.327} &{\scalebox{0.78}{0.193}} &{\scalebox{0.78}{0.296}} &\scalebox{0.78}{0.227} &\scalebox{0.78}{0.338} \\ %&\scalebox{0.78}{0.311} &\scalebox{0.78}{0.397} \\
    \midrule

    \multirow{5}{*}{\rotatebox{90}{\update{\scalebox{0.95}{Exchange}}}}
    &  \scalebox{0.78}{96} & \boldres{\scalebox{0.78}{0.080} }& \boldres{\scalebox{0.78}{0.198} }& \secondres{\scalebox{0.78}{0.086}} & {\scalebox{0.78}{0.206}} & \scalebox{0.78}{0.093} & \scalebox{0.78}{0.217} & {\scalebox{0.78}{0.088}} & \secondres{\scalebox{0.78}{0.205}} & \scalebox{0.78}{0.256} & \scalebox{0.78}{0.367} & \scalebox{0.78}{0.094} & \scalebox{0.78}{0.218} & \scalebox{0.78}{0.107} & \scalebox{0.78}{0.234} & \scalebox{0.78}{0.088} & \scalebox{0.78}{0.218} & \scalebox{0.78}{0.267} & \scalebox{0.78}{0.396} & \scalebox{0.78}{0.148} & \scalebox{0.78}{0.278} & \scalebox{0.78}{0.111} & \scalebox{0.78}{0.237} & \scalebox{0.78}{0.197} & \scalebox{0.78}{0.323} \\ %& \scalebox{0.78}{0.847} & \scalebox{0.78}{0.752} \\
    &  \scalebox{0.78}{192} & \boldres{\scalebox{0.78}{0.173} }& \boldres{\scalebox{0.78}{0.296} }& \scalebox{0.78}{0.177} & {\scalebox{0.78}{0.299}} & \scalebox{0.78}{0.184} & \scalebox{0.78}{0.307} & \secondres{\scalebox{0.78}{0.176}} & \secondres{\scalebox{0.78}{0.299}} & \scalebox{0.78}{0.470} & \scalebox{0.78}{0.509} & \scalebox{0.78}{0.184} & \scalebox{0.78}{0.307} & \scalebox{0.78}{0.226} & \scalebox{0.78}{0.344} & {\scalebox{0.78}{0.176}} & \scalebox{0.78}{0.315} & \scalebox{0.78}{0.351} & \scalebox{0.78}{0.459} & \scalebox{0.78}{0.271} & \scalebox{0.78}{0.315} & \scalebox{0.78}{0.219} & \scalebox{0.78}{0.335} & \scalebox{0.78}{0.300} & \scalebox{0.78}{0.369} \\ %& \scalebox{0.78}{1.204} & \scalebox{0.78}{0.895} \\
    &  \scalebox{0.78}{336} & \secondres{\scalebox{0.78}{0.314} }& \secondres{\scalebox{0.78}{0.406} }& \scalebox{0.78}{0.331} & {\scalebox{0.78}{0.417}} & \scalebox{0.78}{0.351} & \scalebox{0.78}{0.432}& \boldres{\scalebox{0.78}{0.301}} & \boldres{\scalebox{0.78}{0.397}} & \scalebox{0.78}{1.268} & \scalebox{0.78}{0.883} & \scalebox{0.78}{0.349} & \scalebox{0.78}{0.431} & \scalebox{0.78}{0.367} & \scalebox{0.78}{0.448} & {\scalebox{0.78}{0.313}} & \scalebox{0.78}{0.427} & \scalebox{0.78}{1.324} & \scalebox{0.78}{0.853} & \scalebox{0.78}{0.460} & \scalebox{0.78}{0.427} & \scalebox{0.78}{0.421} & \scalebox{0.78}{0.476} & \scalebox{0.78}{0.509} & \scalebox{0.78}{0.524} \\ %& \scalebox{0.78}{1.672} & \scalebox{0.78}{1.036} \\
    &  \scalebox{0.78}{720} & \boldres{\scalebox{0.78}{0.838} }& \secondres{\scalebox{0.78}{0.693} }& {\scalebox{0.78}{0.847}} & \boldres{\scalebox{0.78}{0.691}} & \scalebox{0.78}{0.886} & \scalebox{0.78}{0.714} & \scalebox{0.78}{0.901} & \scalebox{0.78}{0.714} & \scalebox{0.78}{1.767} & \scalebox{0.78}{1.068} & \scalebox{0.78}{0.852} & \scalebox{0.78}{0.698} & \scalebox{0.78}{0.964} & \scalebox{0.78}{0.746} & \secondres{\scalebox{0.78}{0.839}} & \scalebox{0.78}{0.695} & \scalebox{0.78}{1.058} & \scalebox{0.78}{0.797} & \scalebox{0.78}{1.195} & {\scalebox{0.78}{0.695}} & \scalebox{0.78}{1.092} & \scalebox{0.78}{0.769} & \scalebox{0.78}{1.447} & \scalebox{0.78}{0.941} \\ %& \scalebox{0.78}{2.478} & \scalebox{0.78}{1.310} \\
    \cmidrule(lr){2-26}
    &  \scalebox{0.78}{Avg} & \boldres{\scalebox{0.78}{0.351} }& \boldres{\scalebox{0.78}{0.398} }& {\scalebox{0.78}{0.360}} & \secondres{\scalebox{0.78}{0.403}} & \scalebox{0.78}{0.378} & \scalebox{0.78}{0.417} & \scalebox{0.78}{0.367} & {\scalebox{0.78}{0.404}} & \scalebox{0.78}{0.940} & \scalebox{0.78}{0.707} & \scalebox{0.78}{0.370} & \scalebox{0.78}{0.413} & \scalebox{0.78}{0.416} & \scalebox{0.78}{0.443} & \secondres{\scalebox{0.78}{0.354}} & \scalebox{0.78}{0.414} & \scalebox{0.78}{0.750} & \scalebox{0.78}{0.626} & \scalebox{0.78}{0.519} & \scalebox{0.78}{0.429} & \scalebox{0.78}{0.461} & \scalebox{0.78}{0.454} & \scalebox{0.78}{0.613} & \scalebox{0.78}{0.539} \\ %& \scalebox{0.78}{1.550} & \scalebox{0.78}{0.998} \\

    \midrule
    
    \multirow{5}{*}{\rotatebox{90}{\scalebox{0.95}{Traffic}}} 
    & \scalebox{0.78}{96} & \secondres{\scalebox{0.78}{0.402} }& \boldres{\scalebox{0.78}{0.255} }& \boldres{\scalebox{0.78}{0.395}} & \secondres{\scalebox{0.78}{0.268}} & \scalebox{0.78}{0.649} & \scalebox{0.78}{0.389} & {\scalebox{0.78}{0.462}} & \scalebox{0.78}{0.295} & \scalebox{0.78}{0.522} & {\scalebox{0.78}{0.290}} & \scalebox{0.78}{0.805} & \scalebox{0.78}{0.493} &{\scalebox{0.78}{0.593}} &{\scalebox{0.78}{0.321}} &\scalebox{0.78}{0.650} &\scalebox{0.78}{0.396} & \scalebox{0.78}{0.788} & \scalebox{0.78}{0.499} &{\scalebox{0.78}{0.587}} &\scalebox{0.78}{0.366} &\scalebox{0.78}{0.612} &{\scalebox{0.78}{0.338}} &\scalebox{0.78}{0.613} &\scalebox{0.78}{0.388} \\ %&\scalebox{0.78}{0.719} &\scalebox{0.78}{0.391}\\
    & \scalebox{0.78}{192} & \secondres{\scalebox{0.78}{0.426} }& \boldres{\scalebox{0.78}{0.268} }& \boldres{\scalebox{0.78}{0.417}} & \secondres{\scalebox{0.78}{0.276}} & \scalebox{0.78}{0.601} & \scalebox{0.78}{0.366} & {\scalebox{0.78}{0.466}} & \scalebox{0.78}{0.296} & \scalebox{0.78}{0.530} & {\scalebox{0.78}{0.293}} & \scalebox{0.78}{0.756} & \scalebox{0.78}{0.474} &\scalebox{0.78}{0.617} &{\scalebox{0.78}{0.336}} &{\scalebox{0.78}{0.598}} &\scalebox{0.78}{0.370} & \scalebox{0.78}{0.789} & \scalebox{0.78}{0.505} &\scalebox{0.78}{0.604} &\scalebox{0.78}{0.373} &\scalebox{0.78}{0.613} &{\scalebox{0.78}{0.340}} &\scalebox{0.78}{0.616} &\scalebox{0.78}{0.382}  \\ %&\scalebox{0.78}{0.696} &\scalebox{0.78}{0.379} \\
    & \scalebox{0.78}{336} & \secondres{\scalebox{0.78}{0.449} }& \boldres{\scalebox{0.78}{0.275} }& \boldres{\scalebox{0.78}{0.433}} & \secondres{\scalebox{0.78}{0.283}} & \scalebox{0.78}{0.609} & \scalebox{0.78}{0.369} & {\scalebox{0.78}{0.482}} & {\scalebox{0.78}{0.304}} & \scalebox{0.78}{0.558} & \scalebox{0.78}{0.305}  & \scalebox{0.78}{0.762} & \scalebox{0.78}{0.477} &\scalebox{0.78}{0.629} &{\scalebox{0.78}{0.336}}  &{\scalebox{0.78}{0.605}} &\scalebox{0.78}{0.373} & \scalebox{0.78}{0.797} & \scalebox{0.78}{0.508}&\scalebox{0.78}{0.621} &\scalebox{0.78}{0.383} &\scalebox{0.78}{0.618} &{\scalebox{0.78}{0.328}} &\scalebox{0.78}{0.622} &\scalebox{0.78}{0.337} \\ %&\scalebox{0.78}{0.777} &\scalebox{0.78}{0.420} \\
    & \scalebox{0.78}{720} & \secondres{\scalebox{0.78}{0.489} }& \boldres{\scalebox{0.78}{0.297} }& \boldres{\scalebox{0.78}{0.467}} & \secondres{\scalebox{0.78}{0.302}} & \scalebox{0.78}{0.647} & \scalebox{0.78}{0.387} & {\scalebox{0.78}{0.514}} & {\scalebox{0.78}{0.322}} & \scalebox{0.78}{0.589} & \scalebox{0.78}{0.328}  & \scalebox{0.78}{0.719} & \scalebox{0.78}{0.449} &\scalebox{0.78}{0.640} &{\scalebox{0.78}{0.350}} &\scalebox{0.78}{0.645} &\scalebox{0.78}{0.394} & \scalebox{0.78}{0.841} & \scalebox{0.78}{0.523} &{\scalebox{0.78}{0.626}} &\scalebox{0.78}{0.382} &\scalebox{0.78}{0.653} &{\scalebox{0.78}{0.355}} &\scalebox{0.78}{0.660} &\scalebox{0.78}{0.408} \\ %&\scalebox{0.78}{0.864} &\scalebox{0.78}{0.472} \\
    \cmidrule(lr){2-26}
    & \scalebox{0.78}{Avg} & \secondres{\scalebox{0.78}{0.441} }& \boldres{\scalebox{0.78}{0.274} }& \boldres{\scalebox{0.78}{0.428}} & \secondres{\scalebox{0.78}{0.282}} & \scalebox{0.78}{0.626} & \scalebox{0.78}{0.378} & {\scalebox{0.78}{0.481}} & {\scalebox{0.78}{0.304}}& \scalebox{0.78}{0.550} & {\scalebox{0.78}{0.304}} & \scalebox{0.78}{0.760} & \scalebox{0.78}{0.473} &{\scalebox{0.78}{0.620}} &{\scalebox{0.78}{0.336}} &\scalebox{0.78}{0.625} &\scalebox{0.78}{0.383} & \scalebox{0.78}{0.804} & \scalebox{0.78}{0.509} &{\scalebox{0.78}{0.610}} &\scalebox{0.78}{0.376} &\scalebox{0.78}{0.624} &{\scalebox{0.78}{0.340}} &\scalebox{0.78}{0.628} &\scalebox{0.78}{0.379} \\ %&\scalebox{0.78}{0.764} &\scalebox{0.78}{0.416} \\
    \midrule
    
    \multirow{5}{*}{\rotatebox{90}{\scalebox{0.95}{Weather}}} 
    &  \scalebox{0.78}{96} & \boldres{\scalebox{0.78}{0.156} }& \boldres{\scalebox{0.78}{0.202} }& \scalebox{0.78}{0.174} & \secondres{\scalebox{0.78}{0.214}} & \scalebox{0.78}{0.192} & \scalebox{0.78}{0.232} & \scalebox{0.78}{0.177} & {\scalebox{0.78}{0.218}} & \secondres{\scalebox{0.78}{0.158}} & \scalebox{0.78}{0.230}  & \scalebox{0.78}{0.202} & \scalebox{0.78}{0.261} &{\scalebox{0.78}{0.172}} &{\scalebox{0.78}{0.220}} & \scalebox{0.78}{0.196} &\scalebox{0.78}{0.255} & \scalebox{0.78}{0.221} & \scalebox{0.78}{0.306} & \scalebox{0.78}{0.217} &\scalebox{0.78}{0.296} & {\scalebox{0.78}{0.173}} &{\scalebox{0.78}{0.223}} & \scalebox{0.78}{0.266} &\scalebox{0.78}{0.336} \\ %& \scalebox{0.78}{0.300} &\scalebox{0.78}{0.384}  \\
    & \scalebox{0.78}{192} & \secondres{\scalebox{0.78}{0.207} }& \boldres{\scalebox{0.78}{0.250} }& \scalebox{0.78}{0.221} & \secondres{\scalebox{0.78}{0.254}} & \scalebox{0.78}{0.240} & \scalebox{0.78}{0.271} & \scalebox{0.78}{0.225} & \scalebox{0.78}{0.259} & \boldres{\scalebox{0.78}{0.206}} & \scalebox{0.78}{0.277} & \scalebox{0.78}{0.242} & \scalebox{0.78}{0.298} &{\scalebox{0.78}{0.219}} &{\scalebox{0.78}{0.261}}  & \scalebox{0.78}{0.237} &\scalebox{0.78}{0.296} & \scalebox{0.78}{0.261} & \scalebox{0.78}{0.340} & \scalebox{0.78}{0.276} &\scalebox{0.78}{0.336} & \scalebox{0.78}{0.245} &\scalebox{0.78}{0.285} & \scalebox{0.78}{0.307} &\scalebox{0.78}{0.367} \\ %& \scalebox{0.78}{0.598} &\scalebox{0.78}{0.544} \\
    & \scalebox{0.78}{336} & \boldres{\scalebox{0.78}{0.263} }& \boldres{\scalebox{0.78}{0.292} }& {\scalebox{0.78}{0.278}} & \secondres{\scalebox{0.78}{0.296}} & \scalebox{0.78}{0.292} & \scalebox{0.78}{0.307} & \scalebox{0.78}{0.278} & {\scalebox{0.78}{0.297}} & \secondres{\scalebox{0.78}{0.272}} & \scalebox{0.78}{0.335} & \scalebox{0.78}{0.287} & \scalebox{0.78}{0.335} &{\scalebox{0.78}{0.280}} &{\scalebox{0.78}{0.306}} & \scalebox{0.78}{0.283} &\scalebox{0.78}{0.335} & \scalebox{0.78}{0.309} & \scalebox{0.78}{0.378} & \scalebox{0.78}{0.339} &\scalebox{0.78}{0.380} & \scalebox{0.78}{0.321} &\scalebox{0.78}{0.338} & \scalebox{0.78}{0.359} &\scalebox{0.78}{0.395}\\ % &\scalebox{0.78}{0.578} &\scalebox{0.78}{0.523} \\
    & \scalebox{0.78}{720} & \boldres{\scalebox{0.78}{0.340} }& \boldres{\scalebox{0.78}{0.341} }& \scalebox{0.78}{0.358} & \secondres{\scalebox{0.78}{0.347}} & \scalebox{0.78}{0.364} & \scalebox{0.78}{0.353} & \scalebox{0.78}{0.354} & {\scalebox{0.78}{0.348}} & \scalebox{0.78}{0.398} & \scalebox{0.78}{0.418} & {\scalebox{0.78}{0.351}} & \scalebox{0.78}{0.386} &\scalebox{0.78}{0.365} &{\scalebox{0.78}{0.359}} & \secondres{\scalebox{0.78}{0.345}} &{\scalebox{0.78}{0.381}} & \scalebox{0.78}{0.377} & \scalebox{0.78}{0.427} & \scalebox{0.78}{0.403} &\scalebox{0.78}{0.428} & \scalebox{0.78}{0.414} &\scalebox{0.78}{0.410} & \scalebox{0.78}{0.419} &\scalebox{0.78}{0.428} \\ %& \scalebox{0.78}{1.059} &\scalebox{0.78}{0.741} \\
    \cmidrule(lr){2-26}
    & \scalebox{0.78}{Avg} & \boldres{\scalebox{0.78}{0.241} }& \boldres{\scalebox{0.78}{0.271} }& \secondres{\scalebox{0.78}{0.258}} & \secondres{\scalebox{0.78}{0.278}} & \scalebox{0.78}{0.272} & \scalebox{0.78}{0.291} & {\scalebox{0.78}{0.259}} & {\scalebox{0.78}{0.281}} & \scalebox{0.78}{0.259} & \scalebox{0.78}{0.315} & \scalebox{0.78}{0.271} & \scalebox{0.78}{0.320} &{\scalebox{0.78}{0.259}} &{\scalebox{0.78}{0.287}} &\scalebox{0.78}{0.265} &\scalebox{0.78}{0.317} & \scalebox{0.78}{0.292} & \scalebox{0.78}{0.363} &\scalebox{0.78}{0.309} &\scalebox{0.78}{0.360} &\scalebox{0.78}{0.288} &\scalebox{0.78}{0.314} &\scalebox{0.78}{0.338} &\scalebox{0.78}{0.382} \\ %&\scalebox{0.78}{0.634} &\scalebox{0.78}{0.548} \\
    \midrule
    
    \multirow{5}{*}{\rotatebox{90}{\scalebox{0.95}{Solar-Energy}}} 
    &  \scalebox{0.78}{96} & \boldres{\scalebox{0.78}{0.189} }& \boldres{\scalebox{0.78}{0.228} }&\secondres{\scalebox{0.78}{0.203}} &\secondres{\scalebox{0.78}{0.237}} & \scalebox{0.78}{0.322} & \scalebox{0.78}{0.339} & {\scalebox{0.78}{0.234}} & {\scalebox{0.78}{0.286}} &\scalebox{0.78}{0.310} &\scalebox{0.78}{0.331} &\scalebox{0.78}{0.312} &\scalebox{0.78}{0.399} &\scalebox{0.78}{0.250} &\scalebox{0.78}{0.292} &\scalebox{0.78}{0.290} &\scalebox{0.78}{0.378} &\scalebox{0.78}{0.237} &\scalebox{0.78}{0.344} &\scalebox{0.78}{0.242} &\scalebox{0.78}{0.342} &\scalebox{0.78}{0.215} &\scalebox{0.78}{0.249} &\scalebox{0.78}{0.884} &\scalebox{0.78}{0.711}\\ % &\scalebox{0.78}{0.236} &{\scalebox{0.78}{0.259}} \\
    & \scalebox{0.78}{192} & \boldres{\scalebox{0.78}{0.222} }& \boldres{\scalebox{0.78}{0.253} }&\secondres{\scalebox{0.78}{0.233}} &\secondres{\scalebox{0.78}{0.261}} & \scalebox{0.78}{0.359} & \scalebox{0.78}{0.356}& {\scalebox{0.78}{0.267}} & {\scalebox{0.78}{0.310}} &\scalebox{0.78}{0.734} &\scalebox{0.78}{0.725} &\scalebox{0.78}{0.339} &\scalebox{0.78}{0.416} &\scalebox{0.78}{0.296} &\scalebox{0.78}{0.318} &\scalebox{0.78}{0.320} &\scalebox{0.78}{0.398} &\scalebox{0.78}{0.280} &\scalebox{0.78}{0.380} &\scalebox{0.78}{0.285} &\scalebox{0.78}{0.380} &\scalebox{0.78}{0.254} &\scalebox{0.78}{0.272} &\scalebox{0.78}{0.834} &\scalebox{0.78}{0.692} \\ %&\secondres{\scalebox{0.78}{0.217}} &{\scalebox{0.78}{0.269}} \\
    & \scalebox{0.78}{336}& \boldres{\scalebox{0.78}{0.242} }& \secondres{\scalebox{0.78}{0.275}  }&\secondres{\scalebox{0.78}{0.248}} &\boldres{\scalebox{0.78}{0.273}} & \scalebox{0.78}{0.397} & \scalebox{0.78}{0.369}& {\scalebox{0.78}{0.290}}  &{\scalebox{0.78}{0.315}} &\scalebox{0.78}{0.750} &\scalebox{0.78}{0.735} &\scalebox{0.78}{0.368} &\scalebox{0.78}{0.430} &\scalebox{0.78}{0.319} &\scalebox{0.78}{0.330} &\scalebox{0.78}{0.353} &\scalebox{0.78}{0.415} &\scalebox{0.78}{0.304} &\scalebox{0.78}{0.389} &\scalebox{0.78}{0.282} &\scalebox{0.78}{0.376} &\scalebox{0.78}{0.290} &\scalebox{0.78}{0.296} &\scalebox{0.78}{0.941} &\scalebox{0.78}{0.723} \\ %&{\scalebox{0.78}{0.249}} &{\scalebox{0.78}{0.283}}\\
    & \scalebox{0.78}{720} & \boldres{\scalebox{0.78}{0.247} }& \secondres{\scalebox{0.78}{0.282} }&\secondres{\scalebox{0.78}{0.249}} &\boldres{\scalebox{0.78}{0.275}} & \scalebox{0.78}{0.397} & \scalebox{0.78}{0.356} & {\scalebox{0.78}{0.289}} &{\scalebox{0.78}{0.317}} &\scalebox{0.78}{0.769} &\scalebox{0.78}{0.765} &\scalebox{0.78}{0.370} &\scalebox{0.78}{0.425} &\scalebox{0.78}{0.338} &\scalebox{0.78}{0.337} &\scalebox{0.78}{0.356} &\scalebox{0.78}{0.413} &\scalebox{0.78}{0.308} &\scalebox{0.78}{0.388} &\scalebox{0.78}{0.357} &\scalebox{0.78}{0.427} &\scalebox{0.78}{0.285} &\scalebox{0.78}{0.295} &\scalebox{0.78}{0.882} &\scalebox{0.78}{0.717} \\ %&\secondres{\scalebox{0.78}{0.241}} &\secondres{\scalebox{0.78}{0.317}}\\
    \cmidrule(lr){2-26}
    & \scalebox{0.78}{Avg} & \boldres{\scalebox{0.78}{0.225} }& \boldres{\scalebox{0.78}{0.260} }&\secondres{\scalebox{0.78}{0.233}} &\secondres{\scalebox{0.78}{0.262}} & \scalebox{0.78}{0.369} & \scalebox{0.78}{0.356} &{\scalebox{0.78}{0.270}} &{\scalebox{0.78}{0.307}} &\scalebox{0.78}{0.641} &\scalebox{0.78}{0.639} &\scalebox{0.78}{0.347} &\scalebox{0.78}{0.417} &\scalebox{0.78}{0.301} &\scalebox{0.78}{0.319} &\scalebox{0.78}{0.330} &\scalebox{0.78}{0.401} &\scalebox{0.78}{0.282} &\scalebox{0.78}{0.375} &\scalebox{0.78}{0.291} &\scalebox{0.78}{0.381} &\scalebox{0.78}{0.261} &\scalebox{0.78}{0.381} &\scalebox{0.78}{0.885} &\scalebox{0.78}{0.711} \\ %&{\scalebox{0.78}{0.235}} &{\scalebox{0.78}{0.280}}\\
    \midrule
     \multicolumn{2}{c|}{\scalebox{0.78}{{$1^{\text{st}}$ Count}}} & \boldres{\scalebox{0.78}{24} }& \boldres{\scalebox{0.78}{26} }& \scalebox{0.78}{4} & \scalebox{0.78}{{3}} & \scalebox{0.78}{1} & \scalebox{0.78}{{4}} & \scalebox{0.78}{{3}} & \scalebox{0.78}{4} & \scalebox{0.78}{1} & \scalebox{0.78}{0} & \scalebox{0.78}{0} & \scalebox{0.78}{0} & \scalebox{0.78}{0} & \scalebox{0.78}{0} & \scalebox{0.78}{0} & \scalebox{0.78}{0} & \scalebox{0.78}{0} & \scalebox{0.78}{0} & \scalebox{0.78}{3} & \scalebox{0.78}{0} & \scalebox{0.78}{0} & \scalebox{0.78}{0} & \scalebox{0.78}{0} & \scalebox{0.78}{0} \\ %& \scalebox{0.78}{2} & \scalebox{0.78}{1}\\
    \bottomrule
  \end{tabular}
    \end{small}
  \end{threeparttable}
}
\end{table}

\clearpage

% \newpage
% \input{checklist}
\end{document}